\documentclass[11pt]{article}
\usepackage[final]{acl}
\usepackage{times}
\usepackage{latexsym}
\usepackage[T1]{fontenc}
\usepackage[utf8]{inputenc}
\usepackage{microtype}
\usepackage{inconsolata}
\usepackage{url}
\usepackage{booktabs,multirow,graphicx,amsmath,amssymb}
\usepackage{tikz,pgfplots}
\pgfplotsset{compat=1.18}
\usetikzlibrary{positioning,arrows.meta,calc}
\newcommand{\nTA}{18}
\newcommand{\taMean}{78}
\newcommand{\stdMean}{32}
\newcommand{\taTrain}{84}
\newcommand{\stdTrain}{93}
\newcommand{\taWeights}{8,715,983}

\newcommand{\iclMeanFour}{88}

\newcommand{\nNI}{14}
\newcommand{\nNIAll}{18}

\newcommand{\nNIFail}{4}

\newcommand{\niMeanAll}{83}
\newcommand{\nNIPair}{13}
\newcommand{\niPairMean}{93}
\newcommand{\niPairDev}{93}

\newcommand{\niAntonymWeights}{8K}
\newcommand{\niPastWeights}{100.2M}

\newcommand{\niMean}{94}

\newcommand{\niLab}{94}
\newcommand{\niIcl}{90}

\newcommand{\niLabDev}{94}

\newcommand{\niTurnsMed}{92}
\newcommand{\niLabTurnsMed}{86}
\newcommand{\nVerify}{87}
\newcommand{\nKV}{32}
\newcommand{\nVO}{17}
\newcommand{\nOtherC}{38}
\newcommand{\gateKV}{207}
\newcommand{\cosKV}{0.96}
\newcommand{\gateVO}{39}

\newcommand{\gateOther}{15}
\newcommand{\cosOther}{0.45}
\newcommand{\nOtherEmb}{10}

\newcommand{\cosOtherRest}{0.81}

\newcommand{\nAbl}{47}

\newcommand{\nAblOK}{37}
\newcommand{\nAblWhole}{28}

\newcommand{\nAblNeither}{5}
\newcommand{\nAblExc}{10}

\newcommand{\nAblKVRest}{5}

\newcommand{\nActUnits}{23}
\newcommand{\actTask}{6.1}
\newcommand{\actOther}{4.0}
\newcommand{\actText}{0.00}
\newcommand{\nActSelective}{4}

\newcommand{\covMin}{0}
\newcommand{\covMax}{98}
\newcommand{\covMedian}{46}
\newcommand{\nRunsScanned}{69}
\newcommand{\nRunsWithPairs}{34}
\newcommand{\nPairRunsFable}{9 of 15}
\newcommand{\nPairRunsOpus}{9 of 18}
\newcommand{\nPairRunsGemini}{16 of 18}
\newcommand{\nPairRunsAstra}{0 of 18}
\newcommand{\nPairsKept}{911}
\newcommand{\accWithPairs}{90.1}
\newcommand{\nWithPairs}{31}
\newcommand{\accWithoutPairs}{93.0}
\newcommand{\nWithoutPairs}{30}
\newcommand{\nCellsNoPairs}{4}
\newcommand{\nCells}{18}
\newcommand{\shotIclFour}{77.7}
\newcommand{\shotFtFour}{68.1}
\newcommand{\shotLoraFour}{70.4}
\newcommand{\shotRankFour}{77.3}
\newcommand{\shotIclEight}{85.1}

\newcommand{\shotIclSixteen}{89.3}
\newcommand{\shotFtSixteen}{86.3}
\newcommand{\shotLoraSixteen}{87.9}
\newcommand{\shotRankSixteen}{88.4}
\newcommand{\shotIclThirtyTwo}{89.6}
\newcommand{\shotFtThirtyTwo}{90.5}
\newcommand{\shotLoraThirtyTwo}{89.8}
\newcommand{\shotRankThirtyTwo}{88.6}

\newcommand{\shotGainSixteen}{4.1}
\newcommand{\shotGainThirtyTwo}{0.4}
\newcommand{\shotIclOne}{42.4}

\newcommand{\sigMin}{-0.7}
\newcommand{\sigMax}{+6.1}
\newcommand{\sigWorstLo}{3.2}
\newcommand{\sigAugMin}{-3.4}
\newcommand{\sigAugMax}{+4.4}
\newcommand{\sigIclMin}{-1.3}
\newcommand{\sigIclMax}{+3.3}

\newcommand{\strictFableAll}{79.5}

\newcommand{\strictGeminiAll}{83.9}

\newcommand{\strictAstraAll}{75.9}

\newcommand{\strictGeminiThree}{85.8}

\newcommand{\strictGeminiThreeIcl}{89.3}
\newcommand{\strictGeminiThreeFtMin}{86.5}
\newcommand{\strictGeminiThreeFtMax}{88.2}

\newcommand{\strictGeminiThreePMin}{0.062}
\newcommand{\strictGeminiThreePMax}{0.85}

\newcommand{\strictGeminiThreeCells}{18}
\newcommand{\strictGeminiThreeSelfDrop}{5.4}
\newcommand{\strictGeminiThreeSelfP}{<0.001}
\newcommand{\strictGeminiThreeSelfCells}{16}
\newcommand{\strictGeminiThreeBelowMin}{0.7}
\newcommand{\strictGeminiThreeBelowMax}{2.5}

\newcommand{\strictGeminiThreeTightName}{context distillation}
\newcommand{\strictGeminiThreeTightLo}{-4.2}
\newcommand{\strictGeminiThreeTightHi}{-0.1}
\newcommand{\strictGeminiThreeWorstLo}{5.6}
\newcommand{\strictAstraThree}{81.0}

\newcommand{\strictAstraThreePMin}{0.018}
\newcommand{\strictAstraThreePMax}{0.090}
\newcommand{\strictAstraThreeSig}{3}
\newcommand{\strictAstraThreeCells}{18}
\newcommand{\strictAstraThreeSelfDrop}{11.6}
\newcommand{\strictAstraThreeSelfP}{0.001}

\newcommand{\strictAstraThreeBelowMin}{5.5}
\newcommand{\strictAstraThreeBelowMax}{7.2}

\newcommand{\strictAstraThreeTightName}{context distillation}
\newcommand{\strictAstraThreeTightLo}{-12.2}
\newcommand{\strictAstraThreeTightHi}{-2.1}

\newcommand{\nStrictRuns}{42}

\newcommand{\strictFtMin}{85.5}
\newcommand{\strictFtMax}{87.5}
\newcommand{\strictIcl}{88.6}
\newcommand{\nStrictGenRuns}{8}

\newcommand{\largeLoraMean}{73.0}
\newcommand{\largeLoraProg}{89.7}
\newcommand{\largeLoraN}{6}

\newcommand{\largeRankoneMean}{79.0}

\newcommand{\nEditorRuns}{224}

\newcommand{\spend}{779}
\newcommand{\spendLate}{97}
\newcommand{\spendFable}{156}
\newcommand{\spendOpus}{105}
\newcommand{\spendAstra}{95.62}
\newcommand{\spendGemini}{325}
\newcommand{\fableAll}{89.7}
\newcommand{\opusAll}{80.5}
\newcommand{\geminiAll}{87.6}
\newcommand{\astraAll}{92.6}
\newcommand{\errAgreeEdMin}{12}
\newcommand{\errAgreeEdMax}{20}
\newcommand{\errAgreeFtMin}{13}
\newcommand{\errAgreeFtMax}{18}
\newcommand{\cosLastEdMin}{0.67}
\newcommand{\cosLastEdMax}{0.87}
\newcommand{\cosLastFtMin}{0.72}
\newcommand{\cosLastFtMax}{0.82}
\newcommand{\iclMean}{89.3}
\newcommand{\fableMean}{88.5}
\newcommand{\opusMean}{90.4}
\newcommand{\rankoneMean}{88.2}
\newcommand{\fableAgree}{83}
\newcommand{\opusAgree}{85}
\newcommand{\fableCos}{0.70}
\newcommand{\randCos}{0.04}
\newcommand{\loraMean}{87.9}

\newcommand{\ftAugMean}{88.2}
\newcommand{\rankoneAugMean}{89.1}
\newcommand{\loraAugMean}{90.7}

\newcommand{\ftMean}{86.5}
\newcommand{\cdMean}{88.0}

\newcommand{\rankoneWeights}{4,358,143}
\newcommand{\fableQA}{13}
\newcommand{\iclQA}{89}
\newcommand{\rankoneQA}{48}
\newcommand{\fableWeights}{8,512}
\newcommand{\nCoded}{138}
\newcommand{\nTaskVector}{89}
\newcommand{\nGated}{109}
\newcommand{\nNamed}{95}
\newcommand{\nFsZs}{108}

\newcommand{\largeSummary}{Of 14 runs, 12 completed; the completed programs reach a mean of 82.5\% against 88.5\% for 16-shot ICL and 14.4\% zero-shot, with mean agreement 76\% and cosine 0.60; 11 of 14 runs recover at least 80\% of ICL accuracy. The programs are not uniformly sparse: the number of parameters modified ranges from 29K to 7.06B, and the largest modify most of the model while leaving perplexity and ARC-Easy accuracy within 0.9 perplexity points and 8.5 ARC points of the unedited model; their residual-stream shifts are correspondingly less aligned with the ICL shift (cosine as low as 0.07).}
\newcommand{\familiesSummary}{The programs reach 91.0\% against 95.6\% on Qwen3.5-0.8B, 96.9\% against 91.7\% on Llama-3.2-1B and 96.7\% against 99.7\% on Mistral-7B-v0.3 for 16-shot ICL, with 29 of 30 runs completed; agreement and cosine similarity are in the same range as on Qwen2.5 (Table~\ref{tab:families}).}
\newcommand{\cotSummary}{Over 4 runs the updated model answers 59.1\% of held-out problems from the bare prompt, against 40.8\% unedited, 59.0\% with the trigger phrase and 67.4\% with sixteen worked solutions: the updates replace the phrase and close 69\% of the distance to full prompting, and adding the phrase on top of an update changes nothing (58.4\%), as in Appendix~\ref{sec:stack}. Perplexity is unchanged in every run. The most aligned update, a rank-one change to one MLP output projection, induces a residual shift whose cosine similarity with the trigger-induced shift is 0.95 on every layer it reaches.}

\newcommand{\nSmall}{44}

\newcommand{\nMid}{41}
\newcommand{\nMidDmg}{4}
\newcommand{\nBig}{14}
\newcommand{\nBigDmg}{5}

\newcommand{\nIncomplete}{21}
\newcommand{\astraMean}{92.6}
\newcommand{\astraIcl}{89.3}
\newcommand{\gOneFiftyFinished}{16}
\newcommand{\gOneFiftyRuns}{18}
\newcommand{\gOneFiftyMean}{93.0}
\newcommand{\gOneFiftyIcl}{90.1}
\newcommand{\gOneFiftyTurns}{60--134}

\title{Weight-Space Programming Writes In-Context Learning \\ and Matches Fine-Tuning in Language Models}

\author{Konstantinos Skianis \hspace{2em} Aristidis Likas \\
  Department of Computer Science and Engineering, University of Ioannina \\
  \texttt{\{kskianis,arly\}@cs.uoi.gr}}

\begin{document}
\maketitle

\begin{abstract}
Weight-space programming is a frontier language model's recently reported ability to read another network's weights and edit them without gradient descent. We investigate how examples can provide a weight update so that a weaker language model performs the task with nothing in the prompt, and study what the updates are. Our experiments cover six tasks, four frontier editors and three families of weaker language models, without the editors being told anything about in-context learning (ICL). In most runs, the editors measure the ICL task vector, the shift the examples cause in the residual stream at the final prompt token, and write it into the weights, at its sparsest into near-silent MLP units keyed to the prompt format. Results reach the accuracy of ICL and fine-tuning on the same examples, LoRA included, usually leaving perplexity untouched, with worse performance under changes of wording. We also qualify the comparison by counting unfinished runs, observing that one editor falls below fine-tuning; held only to the seen examples, without task pairs of their own, only the strongest editor has similar performance. Given a task and unlabelled inputs, that editor writes its own examples and does nearly as well on unseen items as with labels.
\end{abstract}

\section{Introduction}
Recent work by \citet{takehi2026sparks} shows that a frontier large language model (LLM) placed in a Python kernel with another network's weights can identify what the network computes and act as an editor without gradient descent, a capability called weight-space programming. Independently, the task-vector account of in-context learning (ICL) \citep{hendel2023task,todd2024function} shows that in-context examples act through a nearly query-independent direction in the residual stream, and that adding this direction to a zero-shot forward pass reproduces much of in-context learning.

This paper connects the two works by investigating weight-space programming for writing in-context learning into the weights. An editor model is provided with the language model's weights, sixteen examples with the expected output and the exact zero-shot evaluation format. The model can be run at most fifty times and gradients are not allowed to be computed. The final goal is a model that performs the task zero-shot. Following this motivation, we ask four research questions: (a) does the update work, and (b) how does it compare with fine-tuning? (c) Is it in-context learning, or only similar in performance? (d) Lastly, what did the editor actually write?

Our contributions are as follows: (i) To the best of our knowledge, we present the first controlled evaluation of weight-space programming against prompting and gradient-based methods: six tasks, three splits, four editors, four model sizes and three model families, with scripted and gradient baselines and generalisation, interference and side-effect measurements (Section~\ref{sec:setup}). (ii) We show that the finished runs recover most of ICL's accuracy by modifying 0.002--0.9\% of the parameters (median per editor), are on a par with gradient fine-tuning on the same examples and with $10^2$--$10^5$ times fewer parameters changed than full fine-tuning. This can be attributed in part to the editors' own knowledge. With it, a task statement and unlabelled inputs can replace the labels to within a few points. Without it, the editors perform worse, and taking unfinished runs into consideration, one editor is worse than fine-tuning. In outputs and residual-stream geometry the updates are as close to ICL as fine-tuned models are; they usually leave general language modelling abilities unaffected, but they do not generalise across prompt wordings (Section~\ref{sec:results}). (iii) We describe the updates, or programs, themselves: in \nTaskVector\ of \nCoded\ annotated runs the editor's main strategy was to inject the task vector, usually after naming and measuring it, and \nKV\ of \nVerify\ updates read from the weights are context-gated key-value memory units. A scripted copy of this construction is the most ICL-aligned update of all, but fails where the editors adapt. Applying the readable part of each sparse update alone keeps its accuracy in \nAblOK\ of \nAbl\ runs, in \nAblWhole\ of them because it is the whole update (Section~\ref{sec:mechanism}). (iv) Four extensions are reported in the appendix: updates combined with in-context examples, an update that internalises a reasoning prompt, two tasks in one model, and open-weight editors (Appendices~\ref{sec:stack}--\ref{app:open}). Finally, failures are reported next to successes throughout. Code and results are available online: \url{https://github.com/y3nk0/weight-space-programming-study}.\looseness=-1

\section{Related Work}

\paragraph{Mechanisms of in-context learning.} In-context learning \citep{brown2020gpt3} has been explained as Bayesian inference over a latent task \citep{xie2022bayesian}, as the work of induction heads \citep{olsson2022induction}, and as the implicit execution of a learning algorithm: gradient descent \citep{vonoswald2023transformers,dai2023gpt}, ridge regression \citep{akyurek2023what,bai2023statisticians}, or an implicit low-rank update to the MLP weights induced by the context \citep{dherin2025implicit}. \citet{min2022rethinking} showed that much of the benefit of in-context examples lies in the format and label space they establish. The programs we observe fit this reading: the key detects the format, the value supplies the task.

\paragraph{In-context learning as a vector.} \citet{hendel2023task} and \citet{todd2024function} found that in-context examples compress into a nearly query-independent direction in the residual stream that can be added at one layer of a zero-shot pass, and the latter trace it to a small number of attention heads; \citet{liu2024icv} and \citet{li2025implicit} turned this into inference-time interventions, and \citet{yang2025taskvectors} studied when such vectors emerge in training. All add the vector through a hook. The editors studied here write it into the weights behind a detector that fires only in the task context, without being told that such vectors exist, although most name this literature on their own.

\paragraph{Steering and editing weights.} Difference-of-activations vectors steer behaviour \citep{turner2023actadd,rimsky2024caa} and edit what a model treats as true \citep{hernandez2024remedi}; they are applied unconditionally, and the ungated injection in our scripted baseline is their weight-space analogue. \citet{geva2021kv} read MLP hidden units as key-value pairs, the reading our editors adopt; \citet{meng2022rome} and \citet{meng2023memit} insert facts through closed-form rank-one updates to MLP projections; Transformer-Patcher adds one neuron per correction \citep{huang2023patcher}, and a trigger-gated steering direction can be compiled into the weights as a backdoor \citep{yin2026compiling}; \citet{panigrahi2023skill} show that fine-tuned skills localise to sparse parameter subsets; task arithmetic \citep{ilharco2023task} composes differences of fine-tuned weights. The programs here are such edits with a task specified by examples as the target and an agent as the installer.

\paragraph{Internalising context in the weights.} Context distillation trains a model to reproduce its own prompted outputs \citep{snell2022context}; low-rank adaptation restricts fine-tuning to few parameters \citep{hu2022lora}; zeroth-order methods fine-tune with forward passes only \citep{malladi2023mezo}; \citet{mosbach2023fewshot} compare few-shot fine-tuning with ICL. We use context distillation, full fine-tuning, LoRA and a matched-budget rank-1 update as the gradient-based reference points. Closest to what our editors write, \citet{mazzawi2025transmuting} condense a prompt's implicit update into weight patches and In-Context Compilation \citep{merzin2026icc} writes a context's activation shift into the weights in closed form; we ask instead what an unconstrained editor writes, and we do not include these methods.

\paragraph{Weight-space programming and agents that inspect models.} Recent work by \citet{takehi2026sparks} utilises a frontier LLM and a Python kernel with access to another network's weights. Given a specific budget of forward passes, the authors report that it can read what the network computes, repair a weak model and synthesise small networks. Their editor receives thousands of practice problems to improve an existing ability; ours receives sixteen pairs, or a sentence and unlabelled inputs, to get better at a task, and is compared with prompting and gradient-based alternatives. Concurrent benchmarks score editors on many tasks without prompting or gradient baselines \citep{voncsefalvay2026iwibench,hu2026exploring}. Work by \citet{marasini2026weightbench} states that agents cannot yet patch language models. The setting originates from automated interpretability \citep{bills2023neurons,shaham2024maia}; weight-space programming adds the write operation.

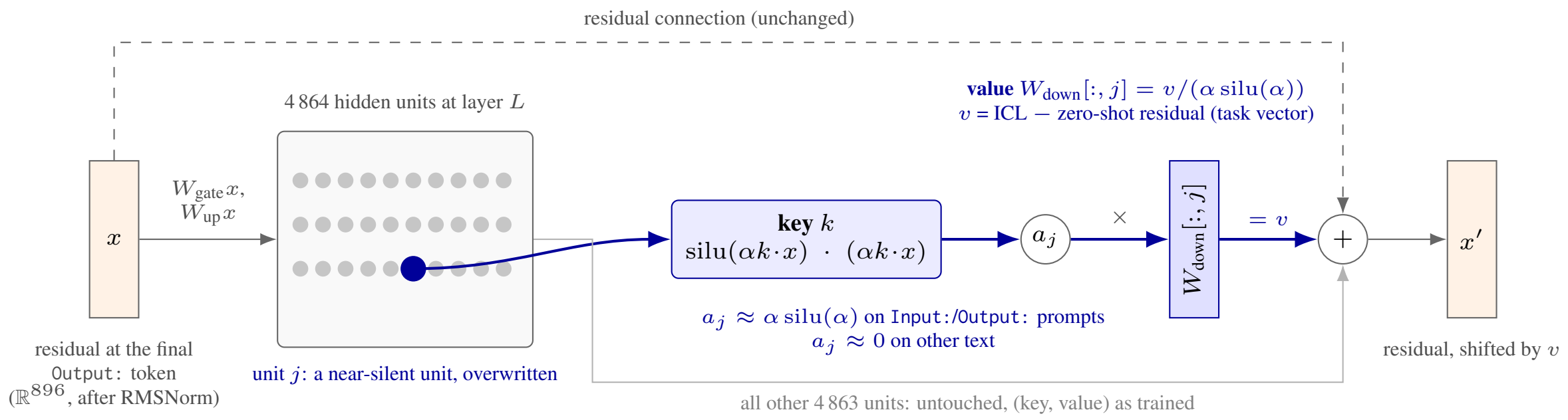
\begin{figure*}[t]\centering
\resizebox{0.98\textwidth}{!}{\begin{tikzpicture}[>=Latex, font=\scriptsize, node distance=6mm,
  vec/.style={draw=black!60, fill=white, minimum width=5mm, minimum height=16mm, inner sep=1pt},
  op/.style={draw=black!60, fill=white, circle, inner sep=1.5pt, minimum size=5mm},
  box/.style={draw=black!60, fill=white, rounded corners=2pt, inner sep=4pt, align=center},
  note/.style={text=black!70, align=left, font=\tiny}]
\node[vec, fill=orange!10] (x) {$x$};
\node[note, below=1mm of x, align=center] {residual at the final\\\texttt{Output:} token\\($\mathbb{R}^{896}$, after RMSNorm)};
\node[box, right=14mm of x, minimum height=22mm, minimum width=26mm, fill=gray!5] (hid) {};
\node[note, above=0.5mm of hid, align=center] {4\,864 hidden units at layer $L$};
\foreach \i in {0,...,9} { \fill[gray!45] ($(hid.west)+(2.4mm+\i*2.3mm, 6mm)$) circle (0.8mm); }
\foreach \i in {0,...,9} { \fill[gray!45] ($(hid.west)+(2.4mm+\i*2.3mm, 1.5mm)$) circle (0.8mm); }
\foreach \i in {0,...,9} { \fill[gray!45] ($(hid.west)+(2.4mm+\i*2.3mm, -3mm)$) circle (0.8mm); }
\fill[blue!60!black] ($(hid.west)+(2.4mm+5*2.3mm, -3mm)$) circle (1.3mm);
\node[note, text=blue!60!black, below=0.5mm of hid, align=center] {unit $j$: a near-silent unit, overwritten};
\node[box, fill=blue!8, draw=blue!60!black, right=14mm of hid] (gate) {\textbf{key} $k$\\ $\mathrm{silu}(\alpha k\!\cdot\!x)\;\cdot\;(\alpha k\!\cdot\!x)$};
\draw[->, blue!60!black, thick] ($(hid.west)+(2.4mm+5*2.3mm, -3mm)$) to[out=0, in=180] (gate.west);
\draw[->, black!60] (x.east) -- node[above, note, align=center] {$W_{\text{gate}}x$,\\$W_{\text{up}}x$} (hid.west);
\node[op, right=8mm of gate] (a) {$a_j$};
\draw[->, blue!60!black, thick] (gate.east) -- (a.west);
\node[note, text=blue!60!black, anchor=north, align=center] at ($(a.south)+(-14.5mm,-3mm)$) {$a_j \approx \alpha\,\mathrm{silu}(\alpha)$ on \texttt{Input:}/\texttt{Output:} prompts\\$a_j \approx 0$ on other text};
\node[vec, fill=blue!12, draw=blue!60!black, right=10mm of a] (v) {\rotatebox{90}{$W_{\text{down}}[:,j]$}};
\node[note, text=blue!60!black, anchor=south east, align=center] at ($(v.east)+(11mm,10.5mm)$) {\textbf{value} $W_{\text{down}}[:,j] = v/(\alpha\,\mathrm{silu}(\alpha))$\\$v$ = ICL $-$ zero-shot residual (task vector)};
\draw[->, blue!60!black, thick] (a.east) -- node[above, note] {$\times$} (v.west);
\node[op, right=10mm of v] (plus) {$+$};
\draw[->, blue!60!black, thick] (v.east) -- node[above, note, text=blue!60!black] {$=v$} (plus.west);
\node[vec, fill=orange!10, right=8mm of plus] (xout) {$x'$};
\draw[->, black!60] (plus.east) -- (xout.west);
\node[note, below=1mm of xout, align=center] {residual, shifted by $v$};
\draw[->, black!60, dashed] (x.north) |- ($(hid.north)+(0,9mm)$) -| (plus.north);
\node[note, text=black!70] at ($(hid.north)+(32mm,9mm)$) [above] {residual connection (unchanged)};
\draw[->, gray!60] (hid.east) -- ++(6mm,0) |- node[pos=0.75, below, note, text=gray] {all other 4\,863 units: untouched, (key, value) as trained} ($(plus.south)+(0,-12mm)$) -- (plus.south);
\end{tikzpicture}}
\caption{The update the editors most often write, drawn for one unit. An MLP hidden unit is a key-value pair \citep{geva2021kv}: its key (row $j$ of $W_\text{gate}$ and $W_\text{up}$) decides when it fires, its value (column $j$ of $W_\text{down}$) is what it adds. The editor writes three vectors of length 896 per unit, with a median of four units per run. Section~\ref{sec:mechanism} gives the weights and checks them against the deltas.}
\label{fig:schematic}
\end{figure*}
\section{Experimental Setup}\label{sec:setup}
\paragraph{Tasks.} We experimented with six word-mapping tasks, already used in the task-vector literature: antonym, country$\to$capital, present$\to$past tense, English$\to$French, $n\to 2n$, and $n\to n{+}7$. On the smallest model, ICL reaches only 81--84\% on the two arithmetic tasks. Each task is split into 16 examples, the only task data provided; 30 development items, used exclusively for baseline hyper-parameter selection; and a test set of 114--190 items. We select three random splits as seeds. Using sixteen examples is a standard few-shot budget \citep{gao2021making} and is where in-context accuracy levels off on these tasks: it gains \shotGainSixteen\ points from 8 to 16 examples and \shotGainThirtyTwo\ from 16 to 32 (Appendix~\ref{app:shots}). For the experiments on larger models (Section~\ref{sec:large}) we add tasks with multi-token, compositional outputs.\looseness=-1
\paragraph{Edited models.} Qwen2.5-0.5B \citep{qwen2025qwen25} in fp32 (24 layers, $d{=}896$, 494M parameters) is the primary edited model. The scale runs use Qwen2.5-1.5B, 7B (fp32) and 14B (bf16).
\paragraph{Evaluation.} A test query $x$ is rendered as \texttt{Input: $x$\textbackslash nOutput:}. The model uses greedy decoding for a task-dependent number of tokens sufficient for the longest reference answer. The output is correct if its first line matches the reference after case folding and accent stripping, with a small set of accepted variants. The reference condition is the unedited model with the same 16 pairs as in-context examples, averaged over three example orders.
\paragraph{Weight-space programming.} The editor is a frontier model (Claude Fable 5.1, Claude Opus 5.5, Gemini 4 Argon or GPT-6 Astra, each at its highest reasoning setting) calling tools on a persistent Python kernel in which the edited model is loaded as a standard PyTorch module. We provide the model's identifier and architecture, the 16 pairs, the evaluation format and metric, and the information that language-modelling degradation will be measured; it may read and overwrite any parameter in place, and code that enables gradients, invokes optimisers or implements training loops is rejected by the harness. The model may be executed at most 50 times, where one execution processes up to 64 sequences and each decoding step counts separately. The editor has 80 turns (150 where stated) and 600\,s per tool call. It receives no information about in-context learning, task vectors or activation steering, and is never shown development or test items. The system prompt is one neutral sentence. The task prompt and the remaining details can be found in Appendices~\ref{app:prompt} and~\ref{app:setup}. Figure~\ref{fig:schematic} shows the update the editors converge on (Section~\ref{sec:mechanism}).
\paragraph{Baselines and controls.} (a) Scripted task vectors, without an LLM: the editors' own construction implemented as a script (a gated key-value unit, in one- and two-unit variants), an ungated constant injection, and an implicit rank-1 update following \citet{dherin2025implicit}, each with layer and scale selected on the dev set. (b) Gradient-based, with Table~\ref{tab:weights} reporting how much of the model every method changes: Adam over all parameters; LoRA \citep{hu2022lora} on the query and value projections of every layer; a rank-1 update to a single MLP matrix (approximately 5.8K trainable parameters, matched to the editors' typical edit size); and context distillation from the ICL model \citep{snell2022context}. Learning rate, step count and layer or rank, when available, are selected on the dev set. The first three are also fitted on the 16 examples plus the examples the editors wrote themselves (Appendix~\ref{app:coverage}). (c) Controls: a random edit of the same tensors with matched Frobenius norm and sparsity pattern, and ICL task-vector patching with the layer selected on the dev set.
\paragraph{Metrics.} We measure zero-shot test accuracy; output agreement, the fraction of test items on which the edited model's zero-shot output equals the ICL model's output; KL divergence from the ICL first-token distribution to the edited model's; for each layer, cosine similarity between the residual-stream shift induced by the edit at the final prompt token and the shift induced by the in-context examples; zero-shot accuracy under three prompt wordings never shown to the editor; 16-shot accuracy on the five other tasks (interference); WikiText-2 perplexity \citep{merity2017pointer}; and ARC-Easy accuracy \citep{clark2018arc}.

\section{Results}\label{sec:results}
\begin{table*}[t]\centering\small
\resizebox{\textwidth}{!}{\begin{tabular}{lrrrrrrr}
\toprule
Method & Antonym & Capital & Past & En$\to$Fr & $2n$ & $n{+}7$ & Mean \\
\midrule
Zero-shot, unedited & 0.0$\pm$0.0 & 1.9$\pm$0.5 & 0.0$\pm$0.0 & 0.0$\pm$0.0 & 0.0$\pm$0.0 & 0.0$\pm$0.0 & 0.3$\pm$0.8 \\
16-shot ICL, unedited & 93.6$\pm$1.4 & 97.0$\pm$0.8 & 93.8$\pm$2.2 & 86.8$\pm$0.4 & 80.6$\pm$3.5 & 83.9$\pm$3.2 & 89.3$\pm$6.4 \\
\midrule
\multicolumn{8}{l}{LLM editors} \\
\quad Fable 5.1 edit & 84.2$\pm$4.7\textsubscript{3} & 87.7$\pm$5.3\textsubscript{3} & 91.4$\pm$1.7\textsubscript{3} & 82.2$\pm$5.9\textsubscript{3} & 99.5$\pm$0.4\textsubscript{3} & 80.6 & 88.5$\pm$7.3 \\
\quad Opus 5.5 edit & 90.5$\pm$3.7\textsubscript{3} & 92.4$\pm$4.3\textsubscript{3} & 90.7$\pm$0.6\textsubscript{2} & 84.2$\pm$0.9\textsubscript{3} & 96.2$\pm$0.5\textsubscript{2} & 91.0 & 90.4$\pm$4.5 \\
\quad Gemini 4 edit (80 turns) & 92.1 & -- & -- & -- & -- & -- & 92.1 \\
\quad Gemini 4 edit (150 turns) & 88.9$\pm$4.2\textsubscript{3} & 93.6$\pm$3.1\textsubscript{3} & 96.8$\pm$0.8\textsubscript{3} & 84.2$\pm$6.3\textsubscript{3} & 99.1$\pm$0.4\textsubscript{3} & 100.0 & 93.0$\pm$6.4 \\
\quad GPT-6 Astra edit & 87.7$\pm$2.7\textsubscript{3} & 93.6$\pm$3.5\textsubscript{3} & 90.1$\pm$2.5\textsubscript{3} & 85.0$\pm$1.6\textsubscript{3} & 99.3$\pm$1.2\textsubscript{3} & 100.0$\pm$0.0\textsubscript{3} & 92.6$\pm$6.1 \\
\multicolumn{8}{l}{Scripted (no LLM)} \\
\quad Scripted gated TV, 1 unit & 83.9$\pm$1.1\textsubscript{3} & 90.4$\pm$1.2\textsubscript{2} & 82.3$\pm$5.7\textsubscript{2} & 37.5$\pm$0.4\textsubscript{3} & 1.4 & 0.0 & 59.2$\pm$34.5 \\
\quad Scripted gated TV, 2 units & 85.3$\pm$3.0\textsubscript{2} & 91.7$\pm$0.6\textsubscript{2} & 83.9$\pm$2.3\textsubscript{2} & 34.9 & 0.7 & 0.0$\pm$0.0\textsubscript{2} & 55.7$\pm$41.6 \\
\quad Scripted constant TV & 85.5$\pm$1.9\textsubscript{2} & 89.5 & 86.3$\pm$0.8\textsubscript{3} & 48.4$\pm$7.1\textsubscript{2} & 0.3$\pm$0.5\textsubscript{2} & 0.7 & 56.2$\pm$38.8 \\
\quad Implicit rank-1 & 11.3$\pm$8.6\textsubscript{2} & 58.8$\pm$5.0\textsubscript{2} & 8.6$\pm$2.8\textsubscript{3} & 0.0$\pm$0.0\textsubscript{2} & 0.0$\pm$0.0\textsubscript{2} & 0.0$\pm$0.0\textsubscript{2} & 12.8$\pm$21.2 \\
\multicolumn{8}{l}{Gradient-based, the 16 pairs} \\
\quad Adam, all weights & 90.5$\pm$1.8\textsubscript{3} & 88.6$\pm$2.3\textsubscript{3} & 93.5$\pm$1.4\textsubscript{3} & 79.3$\pm$1.8\textsubscript{3} & 99.8$\pm$0.4\textsubscript{3} & 67.4$\pm$19.8\textsubscript{3} & 86.5$\pm$12.8 \\
\quad Rank-1 FT, matched & 90.7$\pm$3.2\textsubscript{3} & 87.4$\pm$4.5\textsubscript{3} & 92.5$\pm$1.7\textsubscript{3} & 73.6$\pm$4.1\textsubscript{3} & 98.8$\pm$1.5\textsubscript{3} & 86.3$\pm$1.7\textsubscript{3} & 88.2$\pm$8.3 \\
\quad Context distillation & 90.0$\pm$3.2\textsubscript{3} & 88.3$\pm$1.8\textsubscript{3} & 91.9$\pm$2.4\textsubscript{3} & 79.1$\pm$2.1\textsubscript{3} & 99.3$\pm$1.2\textsubscript{3} & 79.2$\pm$7.8\textsubscript{3} & 88.0$\pm$8.0 \\
\quad LoRA & 90.0$\pm$2.8\textsubscript{3} & 88.0$\pm$2.2\textsubscript{3} & 93.3$\pm$1.2\textsubscript{3} & 75.2$\pm$6.6\textsubscript{3} & 99.1$\pm$1.1\textsubscript{3} & 81.7$\pm$0.4\textsubscript{3} & 87.9$\pm$8.4 \\
\multicolumn{8}{l}{Gradient-based, the 16 pairs and the editors' own} \\
\quad Adam, all weights & 91.4$\pm$0.8\textsubscript{3} & 89.8$\pm$2.5\textsubscript{3} & 95.7$\pm$1.7\textsubscript{3} & 82.7$\pm$3.1\textsubscript{3} & 100.0$\pm$0.0\textsubscript{3} & 69.7$\pm$17.8\textsubscript{3} & 88.2$\pm$11.9 \\
\quad Rank-1 FT, matched & 91.1$\pm$1.9\textsubscript{3} & 86.5$\pm$3.5\textsubscript{3} & 96.0$\pm$0.8\textsubscript{3} & 76.2$\pm$3.7\textsubscript{3} & 99.5$\pm$0.4\textsubscript{3} & 85.2$\pm$1.6\textsubscript{3} & 89.1$\pm$8.1 \\
\quad LoRA & 93.2$\pm$0.9\textsubscript{3} & 90.4$\pm$2.6\textsubscript{3} & 96.0$\pm$1.4\textsubscript{3} & 82.9$\pm$4.8\textsubscript{3} & 99.5$\pm$0.4\textsubscript{3} & 82.4$\pm$2.1\textsubscript{3} & 90.7$\pm$6.9 \\
\bottomrule
\end{tabular}}
\caption{Zero-shot test accuracy (\%) on Qwen2.5-0.5B. Each cell is the mean over the available data splits (seeds); where more than one split was run, the cell shows mean$\pm$standard deviation and a subscript giving the number of splits (e.g.\ 84.2$\pm$4.7\textsubscript{3} is the mean of three runs); cells without a subscript are single runs. The Mean column averages over the runs of the row. All methods are given the same 16 examples; the last block is also given the pairs the editors wrote that lie outside every split (Table~\ref{tab:coverage}) and repeats the block above where there are none. Baselines choose hyper-parameters on 30 dev items. Editor rows include only completed runs; runs that exhausted the turn budget or were interrupted are counted in Table~\ref{tab:editors}.}
\label{tab:headline}
\end{table*}

\paragraph{Accuracy.} Table~\ref{tab:headline} reports zero-shot test accuracy. The unedited model scores 0--3\% on every task, usually by copying the input. The editors' updates reach a mean of \fableMean\% (Fable 5.1) and \opusMean\% (Opus 5.5) against \iclMean\% for 16-shot ICL, and exceed ICL on $2n$ and, for Opus 5.5, on $n{+}7$. These are finished runs. With every run scored as it was left, the four editors reach \fableAll\%, \opusAll\%, \geminiAll\% and \astraAll\% (Table~\ref{tab:editors}), so Opus 5.5 falls below the baselines, which always finish. The scripted implementation of the same construction matches the editors on antonym and capital but fails on translation and arithmetic, and the implicit rank-1 update after \citet{dherin2025implicit} reaches 13\% as implemented here. Both were tuned only over layer and scale. Next, fine-tuning on the same examples, with hyperparameters selected on the development items, does no better than the editors (Table~\ref{tab:weights}): Adam over all parameters reaches \ftMean\% and context distillation \cdMean\%, both changing every weight, LoRA \loraMean\%, and a matched rank-1 update \rankoneMean\% by changing \rankoneWeights\ entries against \fableWeights\ for Fable 5.1. Paired over the splits each editor finished, the editors differ from these four baselines by $\sigMin$ to $\sigMax$ points and from ICL by $\sigIclMin$ to $\sigIclMax$. None of the differences survives correction for multiple comparisons, and no 95\% interval reaches below ${-}\sigWorstLo$ points (Table~\ref{tab:paired} in Appendix~\ref{app:tables}). The comparison flatters the editors in one respect: they know the tasks, and type pairs of their own into their code, correctly labelled and covering \covMin--\covMax\% of the hidden test inputs (Table~\ref{tab:coverage}). Given the editor-written pairs that fall outside every split, Adam, the rank-1 update and LoRA reach \ftAugMean\%, \rankoneAugMean\% and \loraAugMean\%, level with the editors (paired differences of $\sigAugMin$ to $\sigAugMax$ points). When the 16 pairs are the only available task data they may use, Fable 5.1 reaches \strictFableAll\% on the first split (baselines \strictFtMin--\strictFtMax\%); over three splits Gemini 4 Argon reaches \strictGeminiThree\% and GPT-6 Astra \strictAstraThree\% (baselines \strictGeminiThreeFtMin--\strictGeminiThreeFtMax\%), \strictGeminiThreeSelfDrop\ and \strictAstraThreeSelfDrop\ points below their standard runs. Gemini stays within 2.5 points of every baseline; Astra is \strictAstraThreeBelowMin\ to \strictAstraThreeBelowMax\ points below them (Table~\ref{tab:strict}, Appendix~\ref{app:coverage}).\looseness=-1

\begin{table}[t]\centering\small
\resizebox{\columnwidth}{!}{\begin{tabular}{lrrrr}
\toprule
Method & Runs & Entries changed & \% of model & Acc. \\
\midrule
\textbf{LLM editors} & & & & \\
\quad Fable 5.1 edit & 16 & 8,512 & 0.002 & 88.5 \\
\quad Opus 5.5 edit & 14 & 4,358,075 & 0.882 & 90.4 \\
\quad Gemini 4 edit (150 turns) & 16 & 374,080 & 0.076 & 93.0 \\
\quad GPT-6 Astra edit & 18 & 545,664 & 0.110 & 92.6 \\
\quad GPT-6 Astra, wording-agnostic & 18 & 8,715,983 & 1.764 & 83.8 \\
\textbf{Scripted (no LLM)} & & & & \\
\quad Scripted gated TV, 1 unit & 12 & 2,688 & $<$0.001 & 59.2 \\
\quad Scripted gated TV, 2 units & 10 & 5,376 & 0.001 & 55.7 \\
\quad Scripted constant TV & 11 & 896 & $<$0.001 & 56.2 \\
\quad Implicit rank-1 & 13 & 8,716,255 & 1.764 & 12.8 \\
\textbf{Gradient-based} & & & & \\
\quad Rank-1 FT, matched & 18 & 4,358,143 & 0.882 & 88.2 \\
\quad LoRA & 18 & 22,020,064 & 4.457 & 87.9 \\
\quad Context distillation & 18 & 494,015,687 & 99.997 & 88.0 \\
\quad Adam, all weights & 18 & 494,018,107 & 99.997 & 86.5 \\
\textbf{Gradient-based, with the editors' own pairs} & & & & \\
\quad Rank-1 FT, matched & 18 & 4,358,144 & 0.882 & 89.1 \\
\quad LoRA & 18 & 22,020,060 & 4.457 & 90.7 \\
\quad Adam, all weights & 18 & 494,017,854 & 99.997 & 88.2 \\
\bottomrule
\end{tabular}}
\caption{How much of Qwen2.5-0.5B (494M parameters) each method changes, as the median over completed runs, with mean zero-shot accuracy. Entries changed counts parameters whose value differs from the original: a rank-one update touches a whole matrix (about 4.4M entries) while writing roughly 5{,}800 numbers. Adam and context distillation give every parameter a non-zero step. The sparsest editor updates change three vectors per unit.}
\label{tab:weights}
\end{table}
\paragraph{Similarity to in-context learning.} Figure~\ref{fig:cosine} and Table~\ref{tab:icl} in Appendix~\ref{app:tables} compare the updated models with the ICL model. The edited models reproduce the ICL model's output on \fableAgree--\opusAgree\% of test items, against at most 6\% for the unedited zero-shot model; the KL divergence from the ICL first-token distribution decreases by an order of magnitude; and the update's residual shift is aligned with the ICL shift in every layer the update reaches (mean cosine similarity \fableCos\ for Fable 5.1; \randCos\ for the random-edit control). None of this separates the editors from gradient descent. The fine-tuned models agree with ICL on 81 to 83\% of items and are as close in KL. Where ICL is wrong, the edited models repeat its wrong answer on \errAgreeEdMin\ to \errAgreeEdMax\% of items and the fine-tuned ones on \errAgreeFtMin\ to \errAgreeFtMax\%. After the last layer, cosine is \cosLastEdMin\ to \cosLastEdMax\ for the editors and \cosLastFtMin\ to \cosLastFtMax\ for the gradient baselines. The three measures show that an update ends where ICL ends, as an accurate update does; that the editors write ICL's own vector is read from the weights in Section~\ref{sec:mechanism}. Perplexity and ARC-Easy accuracy are unchanged for the Claude editors' updates. The 150-turn Gemini updates reduce 16-shot accuracy on the other tasks by 13 points on average over 16 runs, including two that modify only 4,478 parameters, and two raise perplexity by more than a point. A sparse update can therefore still interfere with other tasks.

\begin{figure}[t]\centering\resizebox{\columnwidth}{!}{\begin{tikzpicture}\begin{axis}[width=0.36\textwidth, height=0.27\textwidth, title={LLM editors}, title style={font=\footnotesize}, xlabel={layer}, ylabel={cosine similarity (ICL shift, edit shift)}, ymin=-0.2, ymax=1, xmin=0, xmax=24, xtick={0,6,12,18,24}, ytick={0,0.5,1}, grid=major, grid style={gray!20}, tick label style={font=\scriptsize}, label style={font=\scriptsize}, legend style={at={(0.03,0.97)}, anchor=north west, font=\tiny, draw=none, fill=none}, legend cell align=left]
\addplot[blue, very thick] coordinates {(0,0.000) (1,0.017) (2,0.068) (3,0.133) (4,0.066) (5,0.070) (6,0.122) (7,0.120) (8,0.111) (9,0.159) (10,0.195) (11,0.187) (12,0.251) (13,0.259) (14,0.338) (15,0.544) (16,0.661) (17,0.754) (18,0.799) (19,0.809) (20,0.822) (21,0.817) (22,0.815) (23,0.807) (24,0.835)};
\addplot[violet, thick] coordinates {(0,0.000) (1,0.009) (2,0.047) (3,0.131) (4,0.106) (5,0.071) (6,0.118) (7,0.145) (8,0.169) (9,0.185) (10,0.206) (11,0.199) (12,0.254) (13,0.251) (14,0.338) (15,0.522) (16,0.605) (17,0.711) (18,0.810) (19,0.826) (20,0.845) (21,0.842) (22,0.840) (23,0.823) (24,0.833)};
\addplot[teal, thick] coordinates {(0,0.000) (1,0.143) (2,0.325) (3,0.430) (4,0.399) (5,0.400) (6,0.446) (7,0.455) (8,0.447) (9,0.501) (10,0.502) (11,0.497) (12,0.526) (13,0.534) (14,0.685) (15,0.736) (16,0.762) (17,0.794) (18,0.829) (19,0.845) (20,0.862) (21,0.865) (22,0.869) (23,0.856) (24,0.868)};
\addplot[red!70!black, thick] coordinates {(0,0.000) (1,0.056) (2,0.078) (3,0.156) (4,0.085) (5,0.080) (6,0.170) (7,0.166) (8,0.164) (9,0.144) (10,0.156) (11,0.150) (12,0.183) (13,0.241) (14,0.236) (15,0.460) (16,0.483) (17,0.527) (18,0.564) (19,0.636) (20,0.653) (21,0.658) (22,0.662) (23,0.648) (24,0.669)};
\addplot[gray, thick, dashed] coordinates {(0,0.000) (1,0.003) (2,0.027) (3,0.058) (4,0.041) (5,0.060) (6,0.074) (7,0.054) (8,0.038) (9,0.027) (10,0.032) (11,0.037) (12,0.063) (13,0.057) (14,0.058) (15,0.056) (16,0.057) (17,0.060) (18,0.055) (19,0.045) (20,0.049) (21,0.038) (22,0.016) (23,0.005) (24,0.056)};
\legend{{Fable 5.1}, {Opus 5.5}, {Gemini 4}, {GPT-6 Astra}, {random edit}}
\end{axis}\end{tikzpicture}
\begin{tikzpicture}\begin{axis}[width=0.36\textwidth, height=0.27\textwidth, title={scripted task vectors}, title style={font=\footnotesize}, xlabel={layer}, ylabel={}, ymin=-0.2, ymax=1, xmin=0, xmax=24, xtick={0,6,12,18,24}, ytick={0,0.5,1}, grid=major, grid style={gray!20}, tick label style={font=\scriptsize}, label style={font=\scriptsize}, legend style={at={(0.03,0.97)}, anchor=north west, font=\tiny, draw=none, fill=none}, legend cell align=left]
\addplot[green!50!black, very thick] coordinates {(0,0.000) (1,0.000) (2,0.000) (3,0.000) (4,0.000) (5,0.000) (6,0.000) (7,0.083) (8,0.072) (9,0.067) (10,0.066) (11,0.051) (12,0.043) (13,0.036) (14,0.191) (15,0.420) (16,0.491) (17,0.571) (18,0.800) (19,0.786) (20,0.872) (21,0.861) (22,0.852) (23,0.835) (24,0.849)};
\addplot[green!50!black, thick, dotted] coordinates {(0,0.000) (1,0.000) (2,0.000) (3,0.000) (4,0.000) (5,0.000) (6,0.000) (7,0.000) (8,0.000) (9,0.089) (10,0.082) (11,0.063) (12,0.060) (13,0.055) (14,0.225) (15,0.292) (16,0.466) (17,0.626) (18,0.877) (19,0.865) (20,0.869) (21,0.855) (22,0.848) (23,0.831) (24,0.840)};
\addplot[green!30!black, thin] coordinates {(0,0.000) (1,0.000) (2,0.000) (3,0.000) (4,0.000) (5,0.000) (6,0.000) (7,0.153) (8,0.066) (9,0.041) (10,0.071) (11,0.059) (12,0.026) (13,0.023) (14,0.112) (15,0.130) (16,0.128) (17,0.158) (18,0.194) (19,0.304) (20,0.319) (21,0.456) (22,0.456) (23,0.418) (24,0.373)};
\addplot[gray, thick, dashed] coordinates {(0,0.000) (1,0.003) (2,0.027) (3,0.058) (4,0.041) (5,0.060) (6,0.074) (7,0.054) (8,0.038) (9,0.027) (10,0.032) (11,0.037) (12,0.063) (13,0.057) (14,0.058) (15,0.056) (16,0.057) (17,0.060) (18,0.055) (19,0.045) (20,0.049) (21,0.038) (22,0.016) (23,0.005) (24,0.056)};
\legend{{gated TV}, {constant TV}, {implicit rank-1}, {random edit}}
\end{axis}\end{tikzpicture}
\begin{tikzpicture}\begin{axis}[width=0.36\textwidth, height=0.27\textwidth, title={gradient baselines}, title style={font=\footnotesize}, xlabel={layer}, ylabel={}, ymin=-0.2, ymax=1, xmin=0, xmax=24, xtick={0,6,12,18,24}, ytick={0,0.5,1}, grid=major, grid style={gray!20}, tick label style={font=\scriptsize}, label style={font=\scriptsize}, legend style={at={(0.03,0.97)}, anchor=north west, font=\tiny, draw=none, fill=none}, legend cell align=left]
\addplot[red!70!black, very thick] coordinates {(0,0.000) (1,0.120) (2,0.052) (3,0.151) (4,0.119) (5,0.078) (6,0.097) (7,0.166) (8,0.242) (9,0.262) (10,0.268) (11,0.266) (12,0.302) (13,0.315) (14,0.407) (15,0.528) (16,0.587) (17,0.645) (18,0.689) (19,0.724) (20,0.754) (21,0.759) (22,0.770) (23,0.747) (24,0.741)};
\addplot[red!70!black, thick, dashed] coordinates {(0,0.000) (1,0.000) (2,0.000) (3,0.000) (4,0.000) (5,0.000) (6,0.000) (7,0.000) (8,0.000) (9,0.035) (10,0.061) (11,0.065) (12,0.090) (13,0.170) (14,0.275) (15,0.412) (16,0.499) (17,0.588) (18,0.636) (19,0.679) (20,0.722) (21,0.738) (22,0.751) (23,0.733) (24,0.749)};
\addplot[orange, thick] coordinates {(0,0.000) (1,-0.266) (2,0.070) (3,0.077) (4,0.049) (5,0.093) (6,0.202) (7,0.258) (8,0.295) (9,0.320) (10,0.337) (11,0.327) (12,0.333) (13,0.329) (14,0.419) (15,0.521) (16,0.585) (17,0.638) (18,0.710) (19,0.750) (20,0.781) (21,0.799) (22,0.805) (23,0.794) (24,0.817)};
\addplot[gray, thick, dashed] coordinates {(0,0.000) (1,0.003) (2,0.027) (3,0.058) (4,0.041) (5,0.060) (6,0.074) (7,0.054) (8,0.038) (9,0.027) (10,0.032) (11,0.037) (12,0.063) (13,0.057) (14,0.058) (15,0.056) (16,0.057) (17,0.060) (18,0.055) (19,0.045) (20,0.049) (21,0.038) (22,0.016) (23,0.005) (24,0.056)};
\legend{{Adam, all}, {rank-1 FT}, {context dist.}, {random edit}}
\end{axis}\end{tikzpicture}}
\caption{Cosine similarity between the residual shift of the update and that of ICL (Section~\ref{sec:setup}), by layer (0 = embedding): mean over completed Qwen2.5-0.5B runs, with the random-edit control for reference.}
\label{fig:cosine}\end{figure}
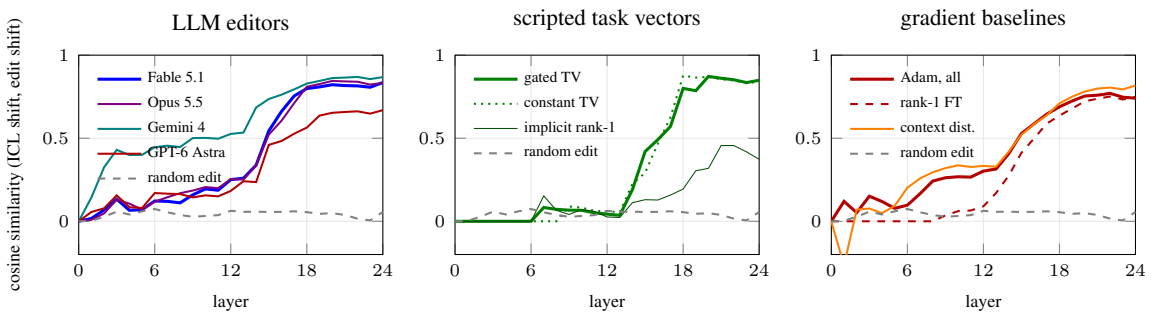
\paragraph{Generalisation across wordings.} The updates do not generalise to prompt wordings that the editor did not see: under a question-answer wording, accuracy falls to \fableQA\% (Fable 5.1), where ICL retains \iclQA\% and the rank-1 gradient update \rankoneQA\%. This is what the gate of Section~\ref{sec:mechanism} predicts: the detector responds to the format it was fitted on. A second condition tells the editor its update will be scored on the mean over four wordings, shown to it, with nothing else changed (Table~\ref{tab:agnostic} in Appendix~\ref{app:tables}). Over \nTA\ runs the mean across wordings rises from \stdMean\% to \taMean\% against \iclMeanFour\% for ICL, while the original wording falls from \stdTrain\% to \taTrain\% and the median update grows to \taWeights\ parameters. The brittleness is therefore mostly the editor optimising what it is scored on: it covers the wordings it is shown, at a cost in precision.

\IfFileExists{tables/nolabel.tex}{\begin{table}[t]\centering\small
\resizebox{\columnwidth}{!}{\begin{tabular}{lccccccc}
\toprule
Condition & Antonym & Capital & Past & En$\to$Fr & $2n$ & $n{+}7$ & Mean \\
\midrule
Unedited, zero-shot & 0.0 & 2.6 & 0.0 & 0.0 & 0.0 & 0.0 & 0.4 \\
16-shot ICL, unedited & 94.2 & 96.5 & 93.8 & 86.6 & 80.0 & 80.6 & 88.6 \\
\midrule
\multicolumn{8}{l}{\textbf{16 labelled examples, test items unseen (5 runs)}} \\
\quad test items & 90.5 & 96.5 & 96.0 & 85.3 & 99.3 & -- & 93.5 \\
\quad held-out items & 96.7 & 96.7 & 93.3 & 83.3 & 100.0 & -- & 94.0 \\
\midrule
\multicolumn{8}{l}{\textbf{No labels, task stated, test inputs shown (6 runs)}} \\
\quad the inputs shown & \textbf{100.0} & \textbf{100.0} & \textbf{100.0} & \textbf{100.0} & \textbf{100.0} & \textbf{100.0} & \textbf{100.0} \\
\quad held-out items & \textbf{6.7} & \textbf{16.7} & \textbf{86.7} & \textbf{0.0} & \textbf{100.0} & \textbf{100.0} & \textbf{51.7} \\
\quad parameters modified & 387K & 54K & 4.4M & 175K & 9.7M & 9.7M & 2.4M \\
\quad perplexity after the update & 17.8 & 17.5 & 17.4 & 18.0 & 33.9 & 752,686 & 17.9 \\
\midrule
\multicolumn{8}{l}{\textbf{No labels, task not stated, test inputs shown (2 runs)}} \\
\quad the inputs shown & 100.0 & -- & 100.0 & -- & -- & -- & 100.0 \\
\quad held-out items & 90.0 & -- & 83.3 & -- & -- & -- & 86.7 \\
\midrule
\multicolumn{8}{l}{\textbf{No labels, task stated, the 16 training inputs only (14 runs)}} \\
\quad test items (unseen) & \textbf{88.4} & \textbf{95.3} & \textbf{96.0} & \textbf{79.1} & \textbf{98.8} & \textbf{100.0} & \textbf{93.9} \\
\quad held-out items & 93.3 & 93.3 & 96.7 & 66.7 & 98.9 & 100.0 & 93.8 \\
\quad 16 labelled examples, same splits & 88.9 & 93.6 & 96.8 & 85.3 & 99.1 & -- & 93.9 \\
\quad parameters modified & 153K & 93K & 33.8M & 758K & 2.4M & 1.2M & 317K \\
\quad perplexity after the update & 17.4 & 17.7 & 18.5 & 17.6 & 19.0 & 22.9 & 17.5 \\
\bottomrule
\end{tabular}}
\caption{Editing with no labelled data: Gemini 4 Argon on Qwen2.5-0.5B, 150 turns; seed 0, except the last block, which averages over 3 data splits and whose labelled row covers the splits on which both runs finished. Held-out: the 30 development items. Unedited perplexity 17.3; the mean column is a median for parameters and perplexity.}
\label{tab:nolabel}
\end{table}}{}
\paragraph{Editing without labelled data.} While fine-tuning needs labels, the editor does not. In a further condition (Table~\ref{tab:nolabel}) the editor, Gemini 4 Argon with 150 turns, receives no input-output pairs: it is given the test inputs with every answer withheld and a one-sentence statement naming the task and the form of the answer. The budget and the rules are those of the labelled runs, and no function in the session scores against a gold answer, so any supervision is the editor's own. On the inputs it was shown every update reaches 100\%, because the editor labels them from its own knowledge and fits the labels. The 30 development items, seen by no editor in any condition, are the rows to compare with the labelled runs, and they split the outcome in two. 
The editor manages to find the rule when the task has one: the two arithmetic updates are digit-level circuits that score 100\% on unseen numbers, and the past-tense update, a linear direction plus memorised irregulars, scores 87\%. Where there is none, on antonym, capital and translation, the editor stores answers item by item and held-out accuracy falls to 0--17\%. The circuits cost about ten million weights each, and perplexity rises to 34 and to 752{,}686. Given the inputs and no statement, the editor inferred antonym and past tense from the lists alone; its updates hold 90\% and 83\% on held-out items. The controlled version shows no test input: the editor gets the same 16 training inputs as the labelled runs with the answers deleted, so that labels are the only difference. Over six tasks and three splits, \nNI\ of \nNIAll\ runs finished within the turn limit and reach \niMean\% on test (\niMeanAll\% with the unfinished runs scored as they stood), against \niIcl\% for ICL. On the \nNIPair\ task-split cells where the labelled editor also finished, the two score \niPairMean\% and \niLab\%, and \niPairDev\% and \niLabDev\% on the development items. The editor makes its own data: on the first antonym split it wrote 64 pairs of its own, disjoint from the inputs, measured the few-shot minus zero-shot residual with them and wrote a \niAntonymWeights\ unit. Without labels it runs out of turns more often, in \nNIFail\ runs against one, all on translation and $n{+}7$. Where both finish, the effort is similar, with a median of \niTurnsMed\ turns against \niLabTurnsMed, though the first past-tense update is a \niPastWeights\ rewrite. Thus we observe that a task statement can stand in for labelled data, since the editor supplies the labels. However, when the test inputs are provided, the editor memorises them wherever the task has no rule.
\begin{table*}[t]\centering\small
\resizebox{\textwidth}{!}{\begin{tabular}{lrrrrrrrrl}
\toprule
Editor & Runs & Finished & Acc. & Acc., all & Agree & Cos.\ sim. & Weights & Turns & Coded strategies (task vector / format rewrite / other; gated) \\
\midrule
Fable 5.1 & 18 & 16 & 88.5$\pm$7.3 & 89.7 & 83.2$\pm$6.6 & 0.70$\pm$0.25 & 8,512 & 51 & 12 / 5 / 1; gated 17/18 \\
Opus 5.5 & 18 & 14 & 90.4$\pm$4.5 & 80.5 & 84.8$\pm$7.0 & 0.67$\pm$0.25 & 4,358,075 & 69 & 10 / 5 / 3; gated 16/18 \\
Gemini 4 (80 turns) & 6 & 1 & 92.1 & 32.5 & 88.4 & 0.89 & 34,944 & 78 & 5 / 0 / 1; gated 4/6 \\
Gemini 4 (150 turns) & 18 & 16 & 93.0$\pm$6.4 & 87.6 & 86.9$\pm$6.6 & 0.71$\pm$0.23 & 374,080 & 98 & 13 / 5 / 0; gated 16/18 \\
GPT-6 Astra & 18 & 18 & 92.6$\pm$6.1 & 92.6 & 84.7$\pm$5.4 & 0.51$\pm$0.27 & 545,664 & 16 & 9 / 8 / 1; gated 10/18 \\
\bottomrule
\end{tabular}}
\caption{Editors on Qwen2.5-0.5B. Finished: runs that ended with a final message and an update in place; the rest exhausted the turn budget or, in one case, were interrupted by the API provider (Appendix~\ref{app:failures}). Accuracy, agreement, cosine similarity, and parameters modified (median) are over finished runs; turns over all runs. Acc., all: every run, with an unfinished run scored as it was left. Strategy annotations follow the taxonomy of Appendix~\ref{app:taxonomy}; gated: updates keyed to the prompt format by a context detector.}
\label{tab:editors}
\end{table*}
\paragraph{Editors.} Table~\ref{tab:editors} compares the four editors. Fable 5.1 completed 16 of 18 runs in a median of about 50 turns with sparse updates (median \fableWeights\ parameters). Opus 5.5 completed 14 of 18 and modified whole matrices with small relative changes (median 4.4M parameters), generalising somewhat better across wordings (31\% under the question-answer wording). Gemini 4 Argon completed only one of six runs within 80 turns. With 150 turns, over three splits, it completed \gOneFiftyFinished\ of \gOneFiftyRuns\ runs in \gOneFiftyTurns\ turns and reached \gOneFiftyMean\% against \gOneFiftyIcl\% for ICL, with updates from 4\,478 parameters to 124M. GPT-6 Astra sits at the other extreme: all 18 runs completed in 4--27 turns at about \$1 per run, against \$4--7 for the others, reaching \astraMean\% against \astraIcl\% for ICL. It writes the same constructions, and its updates are as bound to the wording as theirs. The \nIncomplete\ runs that did not complete are listed in Appendix~\ref{app:failures}.

\paragraph{Scale.} Table~\ref{tab:scale} in Appendix~\ref{app:tables} reports Fable 5.1 on three model sizes under the same budget. The gap to ICL narrows with size. At 7B the updates are within a few points of ICL with at most 0.001\% of parameters modified and no measurable side effects. Generalisation across wordings stays poor, except in part at 1.5B.\looseness=-1

\paragraph{Internalising a prompt.} Providing sixteen worked GSM8K examples and no mention of reasoning, the editors make Qwen2.5-1.5B reason from the bare prompt: 59.1\% against 40.8\% unedited, 59.0\% with the step-by-step trigger appended and 67.4\% with the sixteen examples in context. Perplexity is unchanged, and the most aligned update is a rank-one change to one MLP projection whose residual shift has cosine 0.95 with the trigger's (Appendix~\ref{sec:cot}).

\IfFileExists{tables/large.tex}{\begin{table*}[t]\centering\small
\resizebox{\textwidth}{!}{\begin{tabular}{llccccccc}
\toprule
Model & Condition & antonym & camelCase & Roman & SST-2 & length & last letter & Pig Latin \\
\midrule
Qwen2.5-7B & zero-shot, unedited & 11.6 & 0.0 & 61.7 & 0.0 & 0.0 & 0.0 & 0.0 \\
 & 16-shot ICL, unedited & 99.8 & 99.8 & 98.9 & 96.0 & 85.7 & 68.5 & 43.7 \\
 & \textbf{weight-space program} & \textbf{94.2} & \textbf{90.9} (turn cap) & \textbf{100.0} & \textbf{93.1} & \textbf{100.0} & \textbf{99.3} & \textbf{51.4} \\
 & LoRA on the 16 pairs & 94.2 & 100.0 & 98.7 & 89.7 & 52.8 & 77.5 & 25.4 \\
 & rank-1 update on the 16 pairs & 95.3 & 100.0 & 86.4 & 93.1 & 85.2 & 84.5 & 29.6 \\
 & agreement / cos.\ sim. & 89 / 0.85 & 91 / 0.66 & 99 / 0.85 & 93 / 0.38 & 83 / 0.07 & 68 / 0.17 & 37 / 0.95 \\
 & parameters modified & 43K & 280K & 359.7M & 29K & 7.06B & 2.74B & 6.2M \\ \midrule
Qwen2.5-14B & zero-shot, unedited & 27.9 & 0.0 & 71.4 & 0.0 & 0.0 & 0.7 & 0.0 \\
 & 16-shot ICL, unedited & 100.0 & 100.0 & 100.0 & 96.9 & 98.4 & 90.8 & 73.7 \\
 & \textbf{weight-space program} & \textbf{87.9} & \textbf{96.8} & \textbf{94.8} & \textbf{91.4} & \textbf{16.2} & \textbf{99.3} (turn cap) & \textbf{65.5} \\
 & agreement / cos.\ sim. & 83 / 0.70 & 97 / 0.73 & 95 / 0.57 & 94 / 0.88 & 19 / 0.11 & 93 / 0.21 & 58 / 0.90 \\
 & parameters modified & 123K & 522K & 5.8M & 1.2M & 40.1M & 6.1M & 1.11B \\
\bottomrule
\end{tabular}}
\caption{Larger models and compositional tasks: Gemini 4 Argon (150 turns, seed 0) on Qwen2.5-7B (fp32) and 14B (bf16). Program accuracy with the unedited zero-shot and 16-shot ICL accuracy for reference; agreement and cosine similarity as in Table~\ref{tab:icl}; parenthesised runs did not complete and are reported as left. LoRA and rank-1: fitted on the same 16 pairs with the 0.5B grids and dev-set selection; Adam over all weights and context distillation need about 120\,GB at 7B and were not run.}
\label{tab:large}
\end{table*}}{}
\subsection{Larger Models and Harder Tasks}\label{sec:large}
We extend the study to Qwen2.5-7B and 14B to test whether the findings depend on the small edited model (Table~\ref{tab:large}) with Gemini 4 Argon (150 turns) as editor, on tasks whose outputs are multi-token or compositional (Table~\ref{tab:tasks} in Appendix~\ref{app:setup}). 
\IfFileExists{tables/large.tex}{\largeSummary}{Results are pending at the time of writing.}
At 7B the \largeLoraN\ completed programs average \largeLoraProg\%, against \largeLoraMean\% for LoRA and \largeRankoneMean\% for a rank-1 update on the same 16 pairs. The margin comes from the character-level tasks, where on word length and last letter the editor writes the rule itself, computed from the spelling of every vocabulary token.
Two 14B runs fail: word length reaches 16\% against 98\% for ICL, and last letter hit the turn cap with an ungated update that scores 99\% while raising perplexity from 4.8 to over 10\,000. The accuracy claim therefore holds at scale on the tasks the editor completes, while the sparsity claim does not: beyond 1.5B Gemini 4 Argon writes ever larger updates, the only ones that do damage.

\subsection{Other Model Families}\label{sec:families}
To test whether the findings depend on the Qwen2.5 family, we repeat the six-task protocol of Table~\ref{tab:headline} on three further models (Table~\ref{tab:families} in Appendix~\ref{app:tables}): Llama-3.2-1B \citep{grattafiori2024llama3} and Mistral-7B-v0.3 \citep{jiang2023mistral} with Gemini 4 Argon as editor, and Qwen3.5-0.8B-Base with GPT-6 Astra over three splits. The first two present differences in tokeniser settings, vocabulary and attention layout; the third is a later-generation hybrid with linear attention in eighteen of its twenty-four layers. Only the parameter names given to the editor change. \familiesSummary{}
The update transfers, and so does its failure mode. On Llama, the antonym and arithmetic programs are large (2.2M--55M parameters) and two are damaging: $n\to n{+}7$ reaches 100\% while raising perplexity from 12.9 to 270. The other four leave it within 0.3. Both arithmetic programs beat ICL (100\% against 83.4\% and 80.6\%) at a cosine of 0.01 and 0.00 with the ICL shift. On Mistral every completed program leaves perplexity at 7.7, including a 118M-parameter update. Last, results on the hybrid Qwen3.5 show that every program lands in an MLP projection or the embeddings, never within an attention component.

\section{What the Editors Write}\label{sec:mechanism}

A Qwen2 MLP computes $W_\text{down}\big(\mathrm{silu}(W_\text{gate}x)\odot W_\text{up}x\big)$ over 4\,864 hidden units. The update in Figure~\ref{fig:schematic} sets $W_\text{gate}[j]=W_\text{up}[j]=\alpha k$, where $k$ is a probe for the final token of an \texttt{Input:}/\texttt{Output:} prompt, and $W_\text{down}[:,j]=v/(\alpha\,\mathrm{silu}(\alpha))$, where $v$ is the task vector, so the unit adds exactly $v$ when $k\!\cdot\!x\approx 1$ and about $(k\!\cdot\!x)^2 v\approx 0$ otherwise.

\paragraph{Strategy annotation.} Each run's final explanation and reasoning summaries were annotated with a fixed taxonomy (Appendix~\ref{app:taxonomy}) by a separate model, Claude Sonnet 5, with structured output. Of \nCoded\ annotated runs, \nFsZs\ estimated the few-shot-minus-zero-shot residual, \nTaskVector\ injected a task vector, and \nGated\ gated the update on a context detector. \nNamed\ used the terms task vector or function vector unprompted, so the update is recognised and verified rather than rediscovered. Three details recur that the task-vector papers do not use: a gate, as in neuron-level editing \citep{huang2023patcher,yin2026compiling}, which protects ordinary text (the scripted ungated injection nearly triples perplexity, the gated one leaves it unchanged); a second unit for the discrepancy left by the first; and, in some of Gemini's runs, a projection removing the training words' embedding subspace from the injected vectors.\looseness=-1

\paragraph{Strategy changes.} When the injected vector does not work, the editors change course. In one past-tense run (Fable 5.1) and on English$\to$French at 7B, the editor found a wording the unedited model already answers correctly and made the evaluation format read as that wording: on past tense through a format-token embedding, two layer-0 units and a late-layer unit that steers between inflections, at 7B through five gated layer-3 units. The hidden states still converge on the ICL state in the upper layers. Dropping a failing construction is what separates the editors from the scripted baseline, which averages 59\%.\looseness=-1

\paragraph{What the updates look like in the weights.} The account above is the editors' own, so we also read the deltas directly. Of \nVerify\ completed runs on Qwen2.5-0.5B, \nKV\ write the key-value unit of Figure~\ref{fig:schematic}, \nVO\ change only the values, one whole $W_\text{down}$, and reuse the existing keys, and \nOtherC\ change embedding rows or other matrices. At the deepest edited layer the first two constructions add \gateKV\ and \gateVO\ times more to the residual stream in the task context than on ordinary text (medians), with cosine \cosKV\ to the measured task vector. For the other updates selectivity falls to \gateOther\ times and the cosine to \cosOther\ (\cosOtherRest\ without the \nOtherEmb\ embedding-only rewrites, which the measure reads as zero). Size predicts damage across all edited models: none of the \nSmall\ updates below 100K parameters raises perplexity by more than 5\%, \nMidDmg\ of \nMid\ between 100K and 10M do, and \nBigDmg\ of \nBig\ above 10M do. Small updates stay legible and keep perplexity intact.\looseness=-1

\paragraph{Is the readable part the working one?} For the key-value and value-only updates we apply the part that matches Figure~\ref{fig:schematic} and the rest of the update separately (Table~\ref{tab:ablation}; \nAbl\ runs, without the two of Appendix~\ref{sec:stack}). In \nAblOK\ the readable part alone keeps at least 90\% of the update's accuracy, and in \nAblWhole\ of those it is the entire update. Of the \nAblExc\ exceptions, \nAblNeither\ need both parts and \nAblKVRest, all by Gemini 4 Argon, have a unit present but the accuracy carried by the other changes, so reading a unit in the weights does not show that it does the work. The key is a format detector, not a task detector: the largest units of the \nActUnits\ working key-value runs have a median activation of \actTask\ on their own task, \actOther\ on the other five tasks and \actText\ on ordinary text, and only \nActSelective\ respond twice as strongly to their own task. Two single-task programs rarely stack, while two tasks written jointly match joint fine-tuning (Appendix~\ref{app:twotask}).\looseness=-1

\section{Conclusion}
In this paper, we asked frontier LLMs to edit the weights of weaker models with no gradients. In most runs the editors wrote the ICL task vector, at its sparsest as a context-gated key-value modification of a few MLP units. In outputs and residual geometry the updates are no closer to in-context learning than fine-tuned models are, but the vector can be read in the weights, and they hold across wordings only for those the editor is told it will be scored on. Finished runs match fine-tuning, the sparsest with a few thousand weights, but with the help of what the editors know about the task: when the sixteen examples were used alone, only the strongest stays close to it, and counting unfinished runs one editor falls below. Given a task statement and unlabelled inputs instead of labels, the editor writes its own supervision and reaches \niPairMean\% on unseen items against \niLab\% with labels. At 7B the \largeLoraN\ completed programs average \largeLoraProg\% against \largeLoraMean\% for LoRA and \largeRankoneMean\% for a rank-1 update, but the updates grow, and only the large ones do damage. In our study, weight-space programming adds legibility, not accuracy.\looseness=-1

\typeout{BODYEND page \thepage}\phantomsection\label{bodyend}
\section*{Limitations}
\paragraph{Scope of the evidence.} The main study uses one model family, Qwen2.5; the replication on Llama-3.2-1B and Mistral-7B-v0.3 covers one editor and one split, and the base rates of in-context learning differ across families. All four editors were run with three data splits on the 0.5B model; the scale experiments, the family replication, and the hybrid condition use a single split, so a difference of a few points in those conditions is within what a second split could shift. The tasks are word-mapping and short compositional tasks on which 16-shot ICL is strong; the findings say nothing about tasks the base model cannot perform in context.

\paragraph{What the editor is told.} The editor is provided with the evaluation template and metric, which makes a format-gated update possible and limits its generalisation across wordings. An editor told it would be scored on four wordings generalises much better, at some cost in accuracy (Section~\ref{sec:results}). On five wordings outside that set its updates reach 20.9\%, within the range of the fine-tuned models (17.1--29.8\%) and far below ICL with the examples written in the same wording (79.9\%; Appendix~\ref{app:tables}). The no-label conditions use one editor; those that show the editor the test inputs use one split, and only their 30 held-out items compare with the labelled runs. The editors also bring knowledge of the tasks that no baseline has: about half of the runs type task pairs (examples) of their own, which contain up to \covMax\% of the hidden test inputs (Appendix~\ref{app:coverage}). The strict-data condition that removes this knowledge was run on three splits for Gemini 4 Argon and GPT-6 Astra, on one for Fable 5.1 and not for Opus 5.5, with one run per task, split and editor. For Fable 5.1 the deficit against fine-tuning is a point estimate; for GPT-6 Astra it is significant against \strictAstraThreeSig\ of the four baselines before correction for multiple comparisons (Appendix~\ref{app:coverage}); for Gemini 4 Argon the \strictGeminiThreeCells\ cells show no significant deficit but cannot exclude one of up to \strictGeminiThreeWorstLo\ points (against context distillation the 95\% interval, $-4.2$ to $-0.1$, excludes zero while the sign-flip test gives $p=0.06$); and one per-task drop occurs where the standard run used nothing beyond the 16 pairs either. The \texttt{python} function is the only tool any editor is given, and no provider-side tool is enabled (Appendix~\ref{app:setup}), but the kernel behind it was not sandboxed: it ran on a node with internet access, in the harness's own process, next to the project folder. An audit of every tool call found no network request, no call to another model, and no run that opened a credential file, the task definitions, the results or another run's files. It did find 88 runs importing the harness's helper module and one importing the harness itself, two Opus 5.5 runs listing the project folder, ten searching the file system for word lists and text corpora (WordNet was not installed), and six loading WikiText-2 from the local cache, including the split used here for perplexity. On the same kernel an open-weight editor did read the task definitions and rebuild the test split; that run is set aside, and the kernel was confined for the open-weight runs that followed (Appendix~\ref{app:open}). The forward-pass budget is not an effective constraint on every editor, since Gemini 4 Argon evaluated candidate updates through a CPU re-implementation of the model; budgets stated in forward passes therefore measure different quantities for different editors.

\paragraph{Baselines and measures.} The scripted baselines were tuned only over layer and scale, and the gradient baselines over grids of learning rate, step count and layer or rank, all on 30 development items: realistic for a sixteen-example regime but noisy, and the implicit rank-1 construction in particular may be stronger than our implementation shows. We did not run the closed-form compilers of context into weights \citep{mazzawi2025transmuting,merzin2026icc}, the nearest methods without an agent, nor prompt tuning, nor a zero-shot prompt that states the task in words. The editors were run at 16 examples only; Appendix~\ref{app:shots} varies the count for in-context learning and the gradient baselines, not for the editors. Side effects are assessed with two inexpensive measures, WikiText-2 perplexity and ARC-Easy, rather than a capability suite. Strategy annotations were produced by a language model under a fixed schema and were not validated by human raters.

\paragraph{Reproducibility and cost.} The editors are proprietary models accessed through APIs; their behaviour may change between versions, and two runs were interrupted by provider-side filtering. Editor usage for the study came to \$\spendFable\ for Fable 5.1, about \$\spendOpus\ for Opus 5.5, and about \texteuro300 (\$\spendGemini) for Gemini 4 Argon, as billed by the providers, \$\spendLate\ more at the same rates for the Fable 5.1 and Gemini 4 Argon strict-data runs, made after those bills, and \$\spendAstra\ for GPT-6 Astra computed from the token counts the API reports; approximately \$\spend\ in total for \nEditorRuns\ runs, so the study can be reproduced, but not cheaply.

\section*{AI/LLM Assistance}
The ideas, the experiments, the analysis and the interpretation of the results in this paper are the authors' own. Apart from the models studied here as editors and the model that annotated their strategies (Appendix~\ref{app:taxonomy}), LLM assistants were used to present the results, by producing tables and figures from the experimental outputs and polishing the text, and for minor supporting tasks such as code assistance and checks of the reported numbers against the result files. The authors reviewed everything the assistants produced and take full responsibility for the content.

\section*{Acknowledgements}
We thank Google for providing the credits that enabled us to run the experiments with Gemini 4 Argon and with the two open-weight editors, DeepSeek V4.1 Flash and GLM-5.2, that were run on Google Vertex AI (EDU Credit 423003116).

\bibliography{refs}

\appendix
\section{Experimental Setup in Detail}\label{app:setup}

\subsection{Hardware and Software}
All edited-model computation ran on one shared node with two NVIDIA H200 NVL GPUs, each partitioned into two MIG 3g.71gb instances (four instances of 70\,GB), an AMD EPYC 9335 32-core CPU (128 threads), and 1\,TB of RAM. Each run used a single MIG instance; the 7B model was loaded in fp32 (30\,GB) and the 14B model in bf16 (30\,GB). Software: Python 3.12.3, PyTorch 2.15.0 with CUDA 13.2, Transformers 5.22.1, Datasets 5.1.0, the Anthropic Python SDK 2.5.0, the OpenAI Python SDK 3.46.0, and the Google GenAI SDK 2.34.0. The node was shared with other users' jobs throughout; the forward-pass budgets and all reported accuracies are unaffected by contention, but two runs were lost to out-of-memory errors at load or checkpoint time and were repeated (Appendix~\ref{app:failures}).

\subsection{Edited Models and Decoding}
Qwen2.5-0.5B, 1.5B, 7B, and 14B base models \citep{qwen2025qwen25} from the Hugging Face Hub, in evaluation mode with autograd disabled globally. Tokenisation uses left padding with the end-of-sequence token as pad. Decoding is greedy with a task-dependent length: the number of tokens in the longest reference answer in the task plus two, with a minimum of six (Table~\ref{tab:tasks}). Qwen2.5 tokenises digits individually, so numeric answers of three digits need up to five tokens including the leading space.

\subsection{Scoring}
A generation is normalised by Unicode NFKD decomposition with combining marks removed, lower-casing, truncation at the first newline, and stripping of surrounding quotation marks and trailing punctuation. It is correct if it equals a reference answer, or begins with a reference answer followed by a space or comma. Each task item lists one canonical answer, used as examples, and zero or more accepted variants (for example ``Kiev'' for ``Kyiv'', ``learnt'' for ``learned'').

\subsection{Data Splits}
Items with duplicate inputs are removed; the remaining items are shuffled with a fixed seed and split into the first 16 (examples), the next 30 (development), and the rest (test). Seeds 0, 1, and 2 give three disjoint example sets. The 16 examples are the only task data given to the editor and to every baseline; the editors add pairs from their own knowledge of the tasks (Appendix~\ref{app:coverage}). The development set is used only to select baseline hyper-parameters and the patching layer. In-context learning is evaluated with the same 16 examples, averaged over three orderings.

\begin{table*}[htbp]\centering\small
\resizebox{\textwidth}{!}{\begin{tabular}{lrrrl}
\toprule
Task & Items & Test & Decode & Example pairs (first three of the seed-0 examples) \\
\midrule
Antonym & 236 & 190 & 6 & birth $\to$ death; rare $\to$ common; death $\to$ life \\
Country$\to$capital & 160 & 114 & 8 & Uruguay $\to$ Montevideo; Brazil $\to$ Brasilia; Belarus $\to$ Minsk \\
Present$\to$past & 170 & 124 & 6 & dance $\to$ danced; feel $\to$ felt; smile $\to$ smiled \\
English$\to$French & 175 & 129 & 7 & cold $\to$ froid; school $\to$ école; head $\to$ tête \\
$n\to 2n$ & 189 & 143 & 6 & 177 $\to$ 354; 62 $\to$ 124; 104 $\to$ 208 \\
$n\to n{+}7$ & 190 & 144 & 6 & 177 $\to$ 184; 61 $\to$ 68; 105 $\to$ 112 \\
snake\_case$\to$camelCase & 200 & 154 & 6 & gate\_unit $\to$ gateUnit; big\_write\_last\_train $\to$ bigWriteLastTrain \\
Roman$\to$integer & 200 & 154 & 6 & CXXVII $\to$ 127; CXVIII $\to$ 118; CCIX $\to$ 209 \\
SST-2 sentiment & 220 & 174 & 6 & (sentence) $\to$ positive / negative \\
Word length & 188 & 142 & 6 & cheese $\to$ 6; sweater $\to$ 7; marble $\to$ 6 \\
Last letter & 188 & 142 & 6 & cheese $\to$ e; sweater $\to$ r; marble $\to$ e \\
Pig Latin & 188 & 142 & 7 & cheese $\to$ eesechay; sweater $\to$ eatersway; marble $\to$ arblemay \\
Date$\to$ISO (excluded) & 200 & 154 & 13 & March 5, 2010 $\to$ 2010-03-05; February 6, 1983 $\to$ 1983-02-06 \\
Number words$\to$digits (excluded) & 200 & 154 & 6 & four hundred seventy-nine $\to$ 479; one hundred fifty-one $\to$ 151 \\
Reversed word (excluded) & 188 & 142 & 6 & cheese $\to$ eseehc; sweater $\to$ retaews; marble $\to$ elbram \\
\bottomrule
\end{tabular}}
\caption{Tasks. Each task is split into 16 examples, 30 development items, and the remaining test items (three random splits serve as seeds). Decode: greedy decoding length, set per task to the longest reference answer plus two tokens (minimum 6). Items with duplicate inputs are removed before splitting.}
\label{tab:tasks}
\end{table*}

\paragraph{Tasks for the larger models.} Three tasks built for Section~\ref{sec:large} were dropped: date normalisation and number-word transcription are solved zero-shot by the unedited 7B model (89\% and 100\%), and letter reversal is not solved by 16-shot ICL at either size (14\% and 36\%). On the retained tasks the unedited zero-shot and 16-shot ICL accuracies are those of Table~\ref{tab:large}; Pig Latin (74\% ICL at 14B) is kept as a task where ICL itself is imperfect. In the 7B comparison with LoRA and the rank-1 update, the editor's margin on two of the three character-level tasks comes from writing the rule rather than fitting the examples: on word length it computes the character count of every vocabulary token from its spelling, writes that table into the embeddings, sums it over the word's tokens with an attention head and checks the result against a system word list of 63\,871 words; on last letter it fits a 26-way linear read-out of the final letter on all 68\,923 alphabetic tokens of the vocabulary, labelled from their spelling, and routes it to the output position with one attention head. On Pig Latin, the third, it reproduces the in-context task vectors with gated MLP units, the update used on the small model.

\subsection{The Editor Interface}
The editor interacts through a single tool, \texttt{python}, that executes source code in a persistent kernel and returns standard output, standard error, and the representation of a trailing bare expression, truncated to 16\,000 characters. Each call is limited to 600\,s. The kernel pre-defines: \texttt{model} and \texttt{tok}; \texttt{train\_examples}; \texttt{make\_prompt(x)}, which renders the evaluation format; \texttt{run(prompts, max\_new\_tokens)} for greedy generation; \texttt{eval\_train()}, which scores the training examples in the evaluation format and prints each prediction; \texttt{logits(prompts)} and \texttt{hidden(prompts)}, which return the final-position logits and the final-position residual stream at every layer; \texttt{budget.used} and \texttt{budget.remaining}; and \texttt{torch}, \texttt{numpy}, and \texttt{torch.nn.functional}.

\paragraph{Budget.} A wrapper around the model's forward method counts every call. One call with at most 64 sequences costs one pass; a larger batch costs $\lceil n/64\rceil$ passes; every decoding step inside generation is a separate call. When the budget is exhausted, forward calls raise an exception, but weight edits remain possible. Reading and writing parameters is unmetered.

\paragraph{Forbidden code.} Submitted code is rejected, without execution, if it matches any of: \verb|.backward(|; enabling gradients (\verb|enable_grad|, \verb|set_grad_enabled(True)|, \verb|inference_mode(False)|); \verb|requires_grad=True| or \verb|requires_grad_(True)|; \verb|torch.optim|, \verb|autograd.grad|, \verb|torch.func.grad|, or \verb|.grad|; and any reference to the budget wrapper's internals. Rejections are reported to the editor as a tool result and do not count as turns consumed beyond the call itself.

\paragraph{Editor models and API settings.} Claude Fable 5.1 (\texttt{claude-fable-5-1}) with adaptive thinking at the maximum effort setting and a 32\,000-token output limit; Claude Opus 5.5 (\texttt{claude-opus-5-5}) with the same settings; GPT-6 Astra (\texttt{gpt-6-astra}, Responses API) with reasoning effort set to its highest level, a 32\,000-token output limit, and a hard cost cap of \$40 per run that was never reached; Gemini 4 Argon (\texttt{gemini-4-argon}, Vertex AI, global endpoint) with thinking level set to high, temperature 1.0, a 32\,000-token output limit, and the provider's content-category filters set to their least restrictive level after one run was interrupted at the default setting. Prompt caching is enabled for the Claude editors. Every request declares exactly one tool, the \texttt{python} function above, as a client-side function call; no provider-side tool (web search, hosted code execution, retrieval or grounding) is enabled for any editor, and every tool call the providers returned in the study is a call to that function. The system prompt for all editors is the single sentence ``You are a careful research assistant working in a stateful Python session.'' A run ends when the editor replies without a tool call, when the turn budget (80, or 150 where stated) is exhausted, or when the provider returns no usable response.

\subsection{Task Prompt}\label{app:prompt}
The editor's first and only user message is the following, shown for the antonym task, seed 0, on Qwen2.5-0.5B; the model description, examples, decoding length, and budget are filled in per run. In the hybrid condition the evaluation section instead shows the $k$ fixed in-context examples preceding the query and states the unedited model's accuracy with those examples on the remaining training pairs.

{\footnotesize
\begin{quote}
You have a persistent Python kernel in which a small language model is loaded, plus 16 examples of a task. Your goal: edit the model's weights directly and in place so that the model performs this task when given a query in the format below with NO examples in the prompt. No gradient-based training is allowed; the edits must come from your own analysis and reasoning about the weights and activations.

\textbf{The model.} \texttt{model}: Qwen/Qwen2.5-0.5B (494M parameters, fp32, eval mode, on device \texttt{cuda:0}). Qwen2ForCausalLM; decoder-only transformer with 24 layers, hidden size 896, 14 attention heads (2 KV heads), MLP intermediate size 4864, vocab 151936, tied input/output embeddings = True. Parameter names follow HuggingFace conventions (\texttt{model.embed\_tokens}, \texttt{model.layers[i].\allowbreak self\_attn.\allowbreak\{q,k,v,o\}\_proj}, \texttt{model.layers[i].\allowbreak mlp.\allowbreak\{gate,up,down\}\_proj}, \texttt{model.layers[i].\allowbreak\{input,\allowbreak post\_attention\}\allowbreak\_layernorm}, \texttt{model.norm}, \texttt{lm\_head}). \texttt{tok}: its tokenizer (left padding, pad = eos). Weights are ordinary nn.Parameters. Autograd is globally disabled, so you can read them and write them in place, e.g.\ \texttt{W = model.model.\allowbreak layers[3].\allowbreak mlp.\allowbreak down\_proj.\allowbreak weight; W.data[:, idx] += v}. Edits persist across calls and are what gets evaluated.

\textbf{The task examples (the only task data you get).} \texttt{train\_examples = [("birth", "death"), ("rare", "common"), ("death", "life"), ("possible", "impossible"), ("teach", "learn"), ("late", "early"), ("mother", "father"), ("arrival", "departure"), ("rough", "smooth"), ("pessimistic", "optimistic"), ("new", "old"), ("teacher", "student"), ("summer", "winter"), ("floor", "ceiling"), ("east", "west"), ("brother", "sister")]}

\textbf{How the edited model will be evaluated (hidden test set).} Each held-out test input $x$ (other instances of the same task, not in train\_examples) is formatted exactly as \texttt{Input: \{x\}\textbackslash nOutput:} and the model generates greedily up to 6 new tokens. The output counts as correct if the generated text up to the first newline, stripped and lower-cased, equals the expected answer (minor spelling variants are accepted). The unedited model's zero-shot accuracy in this format is low. The primary metric is this zero-shot test accuracy; we will also report whether the model's general language modelling was damaged, so avoid gratuitous destruction.

\textbf{Rules.} (1) No gradients, backprop, optimizers or training loops (code containing .backward(), grad enabling, torch.optim etc.\ is rejected). (2) Forward-pass budget: the model may be run at most 50 times in total. One forward call with up to 64 sequences counts as 1 pass; a bigger batch costs ceil(n/64) passes. Every decoding step inside generate() is a separate forward call, so generating 4 tokens for 16 prompts costs 4 passes. Reading and editing weights is free and unlimited. Once the budget is exhausted, forward calls raise an error, but you can still edit weights and finish. (3) Each tool call has a 600s time limit and its output is truncated to 16000 characters.

\textbf{Helpers available in the kernel.} \texttt{train\_examples}; \texttt{make\_prompt(x)}; \texttt{run(prompts, max\_new\_tokens=4)}; \texttt{eval\_train(max\_new\_tokens=4)}; \texttt{logits(prompts)}; \texttt{hidden(prompts)}; \texttt{budget.used}, \texttt{budget.remaining}; \texttt{torch}, \texttt{np}, \texttt{F}.

Work step by step. When you are done, reply without a tool call with a concise explanation of what you changed and why.
\end{quote}}

\subsection{Baselines}
\paragraph{Scripted task vectors.} For each candidate layer $L$ in the middle half of the network and scale $s\in\{1.0, 1.3\}$: capture, in three forward passes, the residual stream and MLP activations for the 16 zero-shot prompts, the 16 leave-one-out 15-shot prompts, and eight generic English sentences; set $v = s\cdot\mathrm{mean}(h^{\text{few}}_{L} - h^{\text{zero}}_{L})$ at the final token; fit $k$ by ridge regression (intercept folded into the mean direction) to separate final-token states from all other positions and the generic text; select the unit with the smallest mean $|$activation$|\cdot\|$value$\|$; write $W_\text{gate}[j]=W_\text{up}[j]=\alpha k$ with $\alpha=5$ and $W_\text{down}[:,j]=v/(\alpha\,\mathrm{silu}(\alpha))$. The two-unit variant repeats the procedure at layer $L{+}1$ on the residual discrepancy that remains. The constant variant adds $v/\bar a_j$ to the value column of the unit with the most constant activation, without a key. The implicit rank-1 variant applies $\Delta W = s\,W(a^{\text{few}}-a^{\text{zero}})\,a^{\text{zero}\top}/\|a^{\text{zero}}\|^2$, averaged over the 16 queries, to the gate and up matrices at layer $L$. The configuration with the highest development accuracy is kept.
\paragraph{Gradient-based.} Adam on the training pairs with the loss on answer tokens only; a checkpoint is scored on the development items every 10 steps. Full fine-tuning: learning rates $\{10^{-6}, 3\times10^{-6}, 10^{-5}, 3\times10^{-5}\}$, up to 80 steps. Rank-1: a rank-one update $uv^\top$ to the down-projection of one MLP, at one of five layers between a third and five sixths of the depth ($\{8, 12, 14, 16, 20\}$ on Qwen2.5-0.5B), learning rates $\{3\times10^{-4}, 10^{-3}, 3\times10^{-3}\}$, up to 120 steps. LoRA: rank $\in\{4, 16\}$ with $\alpha=2r$ on the query and value projections of every layer, learning rates $\{10^{-4}, 3\times10^{-4}, 10^{-3}\}$, up to 120 steps. Context distillation: the student, given the zero-shot prompt, minimises the KL divergence to the 16-shot teacher's distribution at every answer token, on the grid of full fine-tuning. The configuration and checkpoint with the highest development accuracy are kept, ties going to the higher accuracy on the 16 pairs.
\paragraph{Controls.} The random-edit control draws Gaussian noise for each modified tensor, matched to the edit's Frobenius norm and restricted to the edit's sparsity pattern. Task-vector patching adds the mean development-set ICL shift at the output of one layer during prefill only, with the layer selected on the development set.

\begin{table*}[htbp]\centering\small
\resizebox{\textwidth}{!}{\begin{tabular}{llrrrrrrrr}
\toprule
Editor and condition & Edited model & Runs & Finished & Turns & Passes & Min/run & Input tok/run & Output tok/run & USD/run \\
\midrule
Fable 5.1, 80 turns & Qwen2.5-0.5B & 18 & 16 & 51 & 42 & 25 & 3.7M & 87K & 6.94 \\
Fable 5.1, 80 turns & Qwen2.5-1.5B & 3 & 3 & 45 & 47 & 19 & 2.4M & 65K & 5.00 \\
Fable 5.1, 80 turns & Qwen2.5-7B & 3 & 3 & 48 & 47 & 19 & 2.6M & 71K & 5.36 \\
Fable 5.1, 80 turns, the 16 pairs only & Qwen2.5-0.5B & 6 & 6 & 46 & 42 & 40 & 3.6M & 84K & 6.79 \\
Opus 5.5, 80 turns & Qwen2.5-0.5B & 18 & 14 & 69 & 41 & 34 & 7.6M & 155K & 5.83 \\
GPT-6 Astra, 80 turns & Qwen2.5-0.5B & 18 & 18 & 16 & 38 & 9 & 0.2M & 15K & 1.15 \\
GPT-6 Astra, 80 turns & Qwen3.5-0.8B-Base & 18 & 18 & 15 & 38 & 8 & 0.2M & 12K & 1.03 \\
GPT-6 Astra, 80 turns & Qwen2.5-1.5B & 3 & 3 & 29 & 45 & 34 & 1.2M & 17K & 2.60 \\
GPT-6 Astra, 80 turns, wording-agnostic & Qwen2.5-0.5B & 18 & 18 & 22 & 49 & 14 & 0.4M & 21K & 1.71 \\
GPT-6 Astra, 80 turns, the 16 pairs only & Qwen2.5-0.5B & 18 & 18 & 15 & 37 & 10 & 0.2M & 13K & 0.99 \\
Gemini 4 Argon, 80 turns & Qwen2.5-0.5B & 7 & 1 & 75 & 10 & 59 & 7.7M & 151K & 2.55 \\
Gemini 4 Argon, 150 turns & Qwen2.5-0.5B & 18 & 16 & 98 & 25 & 107 & 13.1M & 180K & 4.25 \\
Gemini 4 Argon, 150 turns & Qwen2.5-1.5B & 1 & 1 & 143 & 26 & 102 & 27.7M & 186K & 8.69 \\
Gemini 4 Argon, 150 turns & Qwen2.5-7B & 7 & 6 & 97 & 27 & 70 & 13.2M & 176K & 4.27 \\
Gemini 4 Argon, 150 turns & Qwen2.5-14B & 7 & 6 & 98 & 25 & 78 & 13.6M & 188K & 4.41 \\
Gemini 4 Argon, 150 turns & Llama-3.2-1B & 6 & 6 & 85 & 22 & 91 & 9.2M & 150K & 3.00 \\
Gemini 4 Argon, 150 turns & Mistral-7B-v0.3 & 6 & 5 & 86 & 23 & 68 & 11.1M & 161K & 3.59 \\
Gemini 4 Argon, 150 turns, the 16 pairs only & Qwen2.5-0.5B & 18 & 18 & 81 & 21 & 100 & 9.6M & 157K & 3.13 \\
Gemini 4 Argon, 150 turns, hybrid ($k$ examples) & Qwen2.5-0.5B & 5 & 5 & 78 & 28 & 67 & 10.0M & 149K & 3.25 \\
Gemini 4 Argon, 150 turns, no labels & Qwen2.5-0.5B & 26 & 22 & 96 & 26 & 129 & 12.6M & 183K & 4.06 \\
\midrule
All editor runs & & 224 & 203 & & & 202.8\,h total & & & 779 total \\
\bottomrule
\end{tabular}}
\caption{Editor usage and cost by condition. Runs counts every billed run, so the 80-turn row includes one run that the provider's content filter stopped and that was repeated (Appendix~\ref{app:failures}); Table~\ref{tab:editors} leaves it out. Input tokens include cache reads and writes; the prompt cache serves most of the input in every condition (81--98\%). USD/run: for Fable 5.1, Opus 5.5 and Gemini 4 Argon, the totals billed by the providers for the study (\$156; about \$105, of which \$30 was billed for the first six Opus runs and the twelve later runs are counted at the same effective rate; and about \texteuro300 converted at 1.08 USD per EUR) distributed over runs in proportion to each run's token usage; for GPT-6 Astra, computed from the token counts reported by the API. Minutes are wall-clock per run including the editor's API latency; the edited model's forward passes are a negligible part.}
\label{tab:cost}
\end{table*}
\subsection{Measures}
Output agreement, KL divergence, and cosine similarity are computed on the test set with the 16 seed examples. The cosine similarity is taken per layer between the mean-over-items vectors of (ICL state $-$ zero-shot state, unedited weights) and (edited zero-shot state $-$ unedited zero-shot state) at the final prompt token, and is reported averaged over the layers from the first modified tensor onward. Generalisation uses three alternative templates: \texttt{Q: \{x\}\textbackslash nA:}, \texttt{\{x\} ->}, and \texttt{word: \{x\}\textbackslash nanswer:}. Interference is the mean decrease in 16-shot accuracy on the first 60 test items of each of the five other tasks. Perplexity is measured on the first 12\,288 tokens of the WikiText-2 test split in windows of 512 tokens; ARC-Easy accuracy on the first 200 validation questions by summed answer log-likelihood. The weight-level check of Section~\ref{sec:mechanism} covers the \nVerify\ runs on Qwen2.5-0.5B that had finished when it was run: 65 of the main study, 17 of the wording-agnostic condition and 5 of the hybrid condition. Its selectivity is the norm of the mean change of the final-token residual state after the deepest edited layer on 32 test prompts, divided by the same norm on the final tokens of 32 WikiText windows of 64 tokens. Its cosine is taken between the first of these mean changes and the task vector at that layer. Both are reported as medians over runs. The readable-part test leaves out the two hybrid-condition runs, and the unit activations are those of the unit with the largest written value in each run, with its input taken at the MLP at the final prompt token.

\subsection{Additional Tables and Figures}\label{app:tables}
Table~\ref{tab:scale} reports the scale experiment of Section~\ref{sec:results}, Table~\ref{tab:agnostic} the wording-agnostic condition, Table~\ref{tab:families} the model-family replication, Table~\ref{tab:hybrid} the hybrid condition, Table~\ref{tab:paired} the paired comparison of the editors with the baselines and ICL, Table~\ref{tab:ablation} the test of the readable part of each sparse update (Section~\ref{sec:mechanism}), and Figure~\ref{fig:iclcurves} the $k$-shot curves of Appendix~\ref{sec:stack}; all are summarised in the text.

\paragraph{Wordings that no editor was told about.} We also score every finished update on Qwen2.5-0.5B under five further wordings, which we fixed before any result was seen: \texttt{Question: \{x\}\textbackslash nAnswer:}, \texttt{Source: \{x\}\textbackslash nTarget:}, \texttt{x: \{x\}\textbackslash ny:}, \texttt{\{x\} =} and \texttt{The input is \{x\}. The output is}. The updates of GPT-6 Astra told four wordings reach 20.9\% on them, compared to 76.0\% on the three other wordings it was told and 83.8\% on the original one. The editors told one wording reach 4.9--7.3\%. The fine-tuned models reach 29.8\% (LoRA), 24.7\% (Adam), 20.5\% (context distillation) and 17.1\% (rank-1), and ICL with the 16 examples written in each wording reaches 79.9\%. Thus, an update covers the wordings its editor was told about. On other wordings it performs as the fine-tuned models do, far below ICL.
\begin{table*}[t]\centering\small
\resizebox{\textwidth}{!}{\begin{tabular}{lcccccccc}
\toprule
Method & Agree & KL & cos.\ sim. & cos.\ sim.$_\text{rand}$ & Q/A & word/ans. & $\Delta$ICL other & PPL / ARC \\
\midrule
ICL itself, unedited & 100 & 0 & -- & -- & 89.1$\pm$9.3 & 87.9$\pm$11.5 & 0 & 17.34$\pm$0.00 / 66.5$\pm$0.0 \\
\midrule
Fable 5.1 edit\textsubscript{16} & 83.2$\pm$6.6 & 0.34$\pm$0.20 & 0.70$\pm$0.25 & 0.04$\pm$0.09 & 13.4$\pm$25.9 & 21.2$\pm$30.4 & 6.3$\pm$11.1 & 17.38$\pm$0.06 / 66.7$\pm$1.2 \\
Opus 5.5 edit\textsubscript{14} & 84.8$\pm$7.0 & 0.26$\pm$0.16 & 0.67$\pm$0.25 & 0.11$\pm$0.13 & 31.1$\pm$35.4 & 34.0$\pm$36.6 & 5.8$\pm$6.2 & 17.36$\pm$0.03 / 66.8$\pm$1.2 \\
Gemini 4 edit (80 turns)\textsubscript{1} & 88.4 & 0.26 & 0.89 & 0.15 & 1.1 & 20.0 & -0.7 & 17.34 / 67.0 \\
Gemini 4 edit (150 turns)\textsubscript{16} & 86.9$\pm$6.6 & 0.23$\pm$0.16 & 0.71$\pm$0.23 & 0.12$\pm$0.10 & 16.0$\pm$22.7 & 22.9$\pm$32.6 & 12.9$\pm$10.1 & 17.83$\pm$1.19 / 66.7$\pm$2.4 \\
GPT-6 Astra, wording-agnostic\textsubscript{18} & 77.2$\pm$13.2 & 0.48$\pm$0.42 & 0.60$\pm$0.34 & 0.08$\pm$0.12 & 84.7$\pm$12.8 & 84.3$\pm$15.4 & 6.2$\pm$7.4 & 19.93$\pm$4.43 / 64.9$\pm$4.6 \\
GPT-6 Astra edit\textsubscript{18} & 84.7$\pm$5.4 & 0.91$\pm$2.24 & 0.51$\pm$0.27 & 0.15$\pm$0.14 & 12.8$\pm$27.5 & 19.0$\pm$31.6 & 7.3$\pm$5.6 & 17.52$\pm$0.31 / 67.9$\pm$1.7 \\
Scripted gated TV, 1 unit\textsubscript{12} & 58.3$\pm$33.7 & 0.63$\pm$0.47 & 0.87$\pm$0.11 & 0.01$\pm$0.05 & 26.7$\pm$21.6 & 44.8$\pm$31.3 & 0.4$\pm$1.7 & 17.35$\pm$0.01 / 67.8$\pm$0.3 \\
Scripted gated TV, 2 units\textsubscript{10} & 54.2$\pm$40.3 & 0.48$\pm$0.41 & 0.83$\pm$0.16 & 0.01$\pm$0.04 & 26.1$\pm$27.8 & 37.4$\pm$33.8 & 0.4$\pm$1.7 & 17.34 / 67.5 \\
Scripted constant TV\textsubscript{11} & 55.7$\pm$37.9 & 0.41$\pm$0.37 & 0.87$\pm$0.11 & 0.12$\pm$0.06 & 42.2$\pm$34.6 & 41.4$\pm$32.4 & 25.6$\pm$17.0 & 47.14$\pm$28.82 / 58.8$\pm$6.8 \\
Implicit rank-1\textsubscript{13} & 12.4$\pm$20.2 & 3.08$\pm$2.89 & 0.39$\pm$0.20 & 0.00$\pm$0.05 & 13.5$\pm$15.5 & 17.2$\pm$20.5 & 0.9$\pm$1.1 & 17.44$\pm$0.18 / 65.7$\pm$3.5 \\
Adam, all weights\textsubscript{18} & 80.8$\pm$11.8 & 0.30$\pm$0.28 & 0.41$\pm$0.13 & -0.00$\pm$0.02 & 56.0$\pm$30.5 & 62.0$\pm$33.3 & 11.1$\pm$17.7 & 17.48$\pm$0.24 / 66.9$\pm$2.1 \\
Rank-1 FT, matched\textsubscript{18} & 82.7$\pm$6.9 & 0.30$\pm$0.27 & 0.54$\pm$0.10 & 0.04$\pm$0.07 & 48.1$\pm$34.8 & 52.2$\pm$31.6 & 6.9$\pm$10.0 & 17.49$\pm$0.22 / 65.5$\pm$3.8 \\
Context distillation\textsubscript{18} & 82.9$\pm$7.5 & 0.24$\pm$0.20 & 0.42$\pm$0.08 & -0.00$\pm$0.02 & 61.9$\pm$30.4 & 57.0$\pm$26.1 & 7.0$\pm$14.0 & 17.49$\pm$0.29 / 67.5$\pm$1.7 \\
LoRA\textsubscript{18} & 81.0$\pm$7.8 & 0.33$\pm$0.29 & 0.43$\pm$0.10 & 0.14$\pm$0.06 & 59.6$\pm$30.2 & 63.1$\pm$33.0 & 12.6$\pm$16.8 & 18.52$\pm$1.36 / 63.6$\pm$4.9 \\
\bottomrule
\end{tabular}}
\caption{Is the edit in-context learning in the weights? Means over finished Qwen2.5-0.5B runs (subscript: run count). Agree: \% of test items on which the edited model's zero-shot output equals the ICL model's. KL: KL(ICL $\|$ model) over the first answer token (unedited: 4.3--5.5). cos.\ sim.: the residual-shift cosine of Section~\ref{sec:setup}, over the layers the edit reaches; cos.\ sim.$_\text{rand}$: the same for the random-edit control. Q/A, word/ans.: zero-shot accuracy under two wordings not shown to the editor. $\Delta$ICL other: mean drop in 16-shot accuracy on the other five tasks. PPL: WikiText-2 perplexity; ARC: ARC-Easy (seed-0 runs).}
\label{tab:icl}
\end{table*}
\begin{table}[t]\centering\small
\resizebox{\columnwidth}{!}{\begin{tabular}{lccccc}
\toprule
Model & Antonym & Capital & En$\to$Fr & Cos.\ sim. & Q/A \\
\midrule
Qwen2.5-0.5B & 84.2$\pm$4.7 / 93.6$\pm$1.4 & 87.7$\pm$5.3 / 97.0$\pm$0.8 & 82.2$\pm$5.9 / 86.8$\pm$0.4 & 0.70$\pm$0.25 & 13.4$\pm$25.9 \\
Qwen2.5-1.5B & 93.7 / 98.4 & 94.7 / 100.0 & 79.8 / 96.4 & 0.83$\pm$0.06 & 58.6$\pm$33.1 \\
Qwen2.5-7B & 94.2 / 99.8 & 98.2 / 100.0 & 99.2 / 99.2 & 0.76$\pm$0.16 & 1.9$\pm$2.9 \\
\bottomrule
\end{tabular}}
\caption{Model scale (Fable 5.1, seed 0): zero-shot accuracy of the update / 16-shot ICL accuracy of the unedited model, per task; cosine similarity and Q/A-wording generalisation as in Table~\ref{tab:icl}.}
\label{tab:scale}
\end{table}
\IfFileExists{tables/paired.tex}{\begin{table*}[htbp]\centering\small
\resizebox{\textwidth}{!}{\begin{tabular}{lrcccccccc}
\toprule
 & & & \multicolumn{4}{c}{Trained on the 16 pairs} & \multicolumn{3}{c}{Trained on the 16 pairs and the editors' own} \\ \cmidrule(lr){4-7}\cmidrule(lr){8-10}
Editor & Splits & 16-shot ICL & Adam & LoRA & Context dist. & Rank-1 & Adam & Rank-1 & LoRA \\
\midrule
Fable 5.1 & 16 & $-1.3$ {\scriptsize[$-6.2$, $+4.4$]} & $-0.7$ {\scriptsize[$-3.2$, $+1.7$]} & $-0.1$ {\scriptsize[$-2.8$, $+2.7$]} & $+0.0$ {\scriptsize[$-2.5$, $+2.4$]} & $+0.1$ {\scriptsize[$-3.0$, $+3.3$]} & $-2.3$ {\scriptsize[$-4.5$, $-0.3$]}$^{\ast}$ & $-1.2$ {\scriptsize[$-4.0$, $+1.7$]} & $-3.4$ {\scriptsize[$-5.6$, $-1.6$]}$^{\ast}$ \\
Opus 5.5 & 14 & $+0.0$ {\scriptsize[$-3.4$, $+3.9$]} & $+2.4$ {\scriptsize[$-0.3$, $+5.8$]} & $+3.0$ {\scriptsize[$+0.4$, $+5.8$]} & $+3.0$ {\scriptsize[$+0.3$, $+6.4$]} & $+3.3$ {\scriptsize[$+0.4$, $+6.5$]} & $+0.8$ {\scriptsize[$-1.8$, $+3.9$]} & $+2.1$ {\scriptsize[$-1.0$, $+5.3$]} & $-0.5$ {\scriptsize[$-2.7$, $+1.7$]} \\
Gemini 4 Argon & 16 & $+2.8$ {\scriptsize[$-1.5$, $+7.6$]} & $+5.4$ {\scriptsize[$+0.8$, $+12.7$]}$^{\ast}$ & $+4.3$ {\scriptsize[$+1.8$, $+7.3$]}$^{\ast}$ & $+3.7$ {\scriptsize[$+1.4$, $+6.2$]}$^{\ast}$ & $+4.4$ {\scriptsize[$+1.8$, $+7.1$]}$^{\ast}$ & $+3.7$ {\scriptsize[$-0.6$, $+10.6$]} & $+3.5$ {\scriptsize[$+0.9$, $+6.4$]}$^{\ast}$ & $+1.3$ {\scriptsize[$-1.2$, $+4.5$]} \\
GPT-6 Astra & 18 & $+3.3$ {\scriptsize[$-1.3$, $+8.3$]} & $+6.1$ {\scriptsize[$+0.5$, $+13.3$]} & $+4.7$ {\scriptsize[$+1.1$, $+8.6$]}$^{\ast}$ & $+4.6$ {\scriptsize[$+1.0$, $+8.9$]}$^{\ast}$ & $+4.4$ {\scriptsize[$+1.1$, $+7.7$]}$^{\ast}$ & $+4.4$ {\scriptsize[$-0.9$, $+11.3$]} & $+3.5$ {\scriptsize[$-0.1$, $+7.1$]} & $+1.9$ {\scriptsize[$-1.7$, $+5.8$]} \\
\midrule
Mean of the editors & 18 & $+2.7$ {\scriptsize[$-1.6$, $+7.3$]} & $+5.4$ {\scriptsize[$+0.5$, $+12.4$]} & $+4.1$ {\scriptsize[$+1.2$, $+7.3$]}$^{\ast}$ & $+4.0$ {\scriptsize[$+1.1$, $+7.3$]}$^{\ast}$ & $+3.7$ {\scriptsize[$+1.0$, $+6.5$]}$^{\ast}$ & $+3.7$ {\scriptsize[$-0.9$, $+10.4$]} & $+2.8$ {\scriptsize[$-0.1$, $+6.0$]} & $+1.2$ {\scriptsize[$-1.7$, $+4.7$]} \\
\bottomrule
\end{tabular}}
\caption{Paired comparison on Qwen2.5-0.5B: mean difference in zero-shot test accuracy (points), editor minus reference, over the data splits the editor finished, with the reference taken on the same splits; in brackets a 95\% bootstrap interval over splits (20{,}000 resamples). $^{\ast}$: $p<0.05$ in an exact sign-flip test on the paired differences, uncorrected; none of the 40 comparisons shown stays below 0.05 after Holm correction. The last row averages, on each split, the editors that finished it. The three splits of a task draw on one item pool, so the units are not fully independent.}
\label{tab:paired}
\end{table*}}{}
\IfFileExists{tables/ablation.tex}{\begin{table}[t]\centering\small
\resizebox{\columnwidth}{!}{\begin{tabular}{lrrrrrrr}
\toprule
 & & \multicolumn{4}{c}{Accuracy is carried by} & \multicolumn{2}{c}{Accuracy of} \\
\cmidrule(lr){3-6}\cmidrule(lr){7-8}
Editor & Runs & whole & part & rest & neither & update & part \\
\midrule
\multicolumn{8}{l}{\textbf{Key--value units}} \\
\quad Fable 5.1 & 14 & 13 & 0 & 0 & 1 & 88.3 & 81.9 \\
\quad Opus 5.5 & 5 & 3 & 0 & 0 & 2 & 91.1 & 56.5 \\
\quad Gemini 4 Argon & 12 & 4 & 3 & 5 & 0 & 91.9 & 52.7 \\
\multicolumn{8}{l}{\textbf{Value-only updates}} \\
\quad Opus 5.5 & 4 & 2 & 1 & 0 & 1 & 89.4 & 83.5 \\
\quad GPT-6 Astra & 12 & 6 & 5 & 0 & 1 & 88.1 & 78.7 \\
\midrule
All & 47 & 28 & 9 & 5 & 5 & 89.5 & 71.1 \\
\bottomrule
\end{tabular}}
\caption{Is the part of an update that matches Figure~\ref{fig:schematic} the part that works? For every run classified as a key--value unit or a value-only update, except the two of the hybrid condition (Appendix~\ref{sec:stack}), the readable part (the unit's rows and columns, or the leading singular direction of the output-matrix change) and the rest of the update are applied separately and scored on the test set. Whole: the update is nothing but the readable part. Part: other weights were changed too, and the readable part alone keeps at least 90\% of the accuracy. Rest: the rest alone does. Neither: both are needed. The last two columns give the mean accuracy of the full update and of its readable part alone.}
\label{tab:ablation}
\end{table}}{}
\IfFileExists{tables/agnostic.tex}{\begin{table}[t]\centering\small
\resizebox{\columnwidth}{!}{\begin{tabular}{lccccc}
\toprule
Condition & Input/Output & Q/A & arrow & word/answer & Mean \\
\midrule
16-shot ICL, unedited & 89.3$\pm$6.4 & 88.6$\pm$8.9 & 86.0$\pm$14.4 & 87.7$\pm$10.8 & 87.9$\pm$9.6 \\
\midrule
GPT-6 Astra, told one wording (18 runs) & 92.6$\pm$6.1 & 12.8$\pm$27.5 & 4.7$\pm$10.0 & 19.0$\pm$31.6 & 32.3$\pm$16.7 \\
\textbf{GPT-6 Astra, told four (18 runs)} & \textbf{83.8$\pm$16.0} & \textbf{84.7$\pm$12.8} & \textbf{59.0$\pm$32.4} & \textbf{84.3$\pm$15.4} & \textbf{77.9$\pm$18.3} \\
\midrule
Parameters modified (median) & \multicolumn{2}{c}{545,664 (one)} & \multicolumn{3}{c}{8,715,983 (four)} \\
Perplexity after the update & \multicolumn{2}{c}{17.52$\pm$0.31} & \multicolumn{3}{c}{19.93$\pm$4.43} \\
\bottomrule
\end{tabular}}
\caption{Telling the editor what it will be scored on. Both conditions are GPT-6 Astra on Qwen2.5-0.5B over three splits of the six tasks, with the same examples, budget and rules; they differ only in the evaluation described in the prompt. The first is scored on the \texttt{Input:}/\texttt{Output:} wording alone, the second on the mean over all four wordings shown here, which it is given. Columns report zero-shot test accuracy under each wording; the unedited model scores 0--3\% in all of them. Perplexity of the unedited model is 17.3.}
\label{tab:agnostic}
\end{table}}{}
\IfFileExists{tables/hybrid.tex}{\begin{table}[t]\centering\small
\resizebox{\columnwidth}{!}{\begin{tabular}{lcccccc}
\toprule
Task & $k$ & Base, $k$ examples & Edit$_0$ + $k$ examples & Hybrid edit & ICL-16 & PPL / ARC \\
\midrule
Antonym & 2 & 68.9 & 89.8 & \textbf{92.1} & 94.2 & 17.8 / 63.5 \\
Antonym & 4 & 85.3 & 91.4 & \textbf{91.6} & 94.2 & 17.6 / 68.0 \\
Antonym & 8 & 92.1 & 93.0 & \textbf{95.8} & 94.2 & 18.2 / 61.0 \\
Capital & 4 & 98.2 & -- & \textbf{100.0} & 96.5 & 17.4 / 67.0 \\
Past & 4 & 94.4 & -- & \textbf{95.2} & 93.8 & 17.7 / 66.0 \\
\bottomrule
\end{tabular}}
\caption{Hybrid condition (Gemini 4 Argon, 150 turns, seed 0): the editor is informed that every test prompt will contain the same $k$ in-context examples (the first $k$ training pairs) and optimises the update for that setting. Base: unedited model with those $k$ in-context examples. Edit$_0$: the zero-shot update for the same task evaluated with $k$ in-context examples. ICL-16: unedited model with all 16 examples. Unedited perplexity 17.3, ARC-Easy 66.5.}
\label{tab:hybrid}
\end{table}}{}
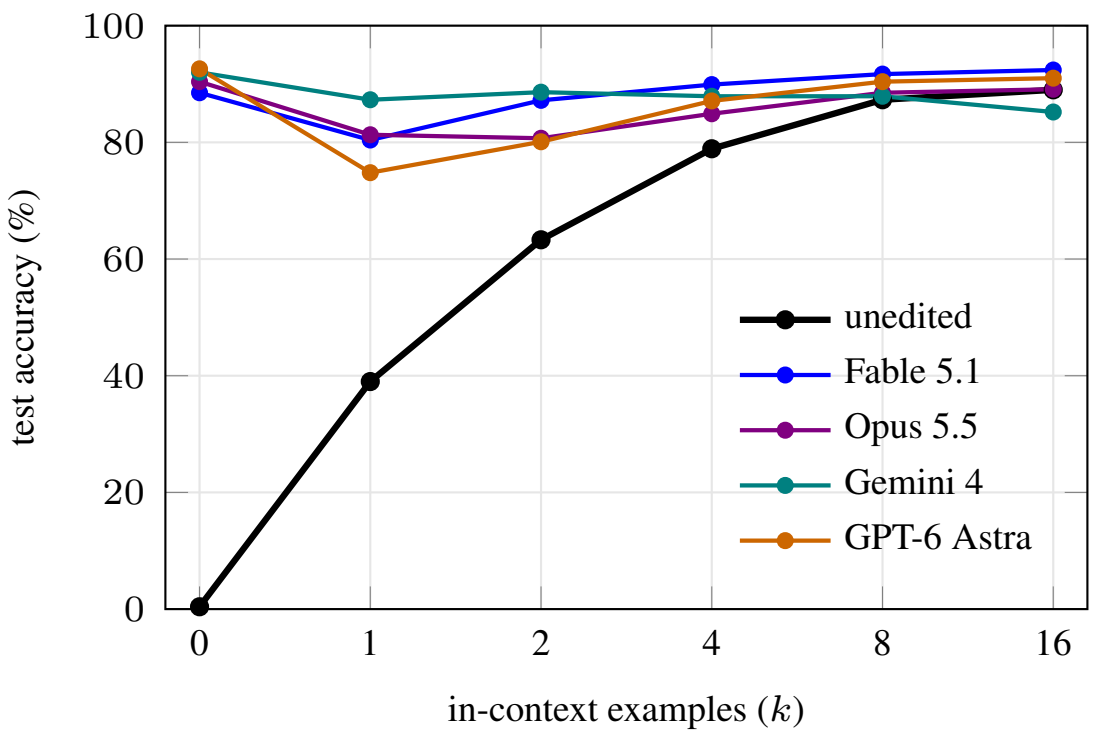
\begin{figure}[t]\centering\begin{tikzpicture}\begin{axis}[width=0.48\textwidth, height=0.34\textwidth, xlabel={in-context examples ($k$)}, ylabel={test accuracy (\%)}, xtick={0,1,2,3,4,5}, xticklabels={0,1,2,4,8,16}, xmin=-0.2, xmax=5.2, ymin=0, ymax=100, grid=major, grid style={gray!20}, tick label style={font=\scriptsize}, label style={font=\scriptsize}, legend style={at={(0.97,0.05)}, anchor=south east, font=\scriptsize, draw=none, fill=none}, legend cell align=left]
\addplot[black, very thick, mark=*, mark size=1.2pt] coordinates {(0,0.4) (1,39.0) (2,63.3) (3,78.9) (4,87.3) (5,89.0)};
\addplot[blue, thick, mark=*, mark size=1.2pt] coordinates {(0,88.5) (1,80.4) (2,87.2) (3,89.9) (4,91.7) (5,92.4)};
\addplot[violet, thick, mark=*, mark size=1.2pt] coordinates {(0,90.4) (1,81.3) (2,80.7) (3,84.9) (4,88.5) (5,89.1)};
\addplot[teal, thick, mark=*, mark size=1.2pt] coordinates {(0,92.0) (1,87.3) (2,88.6) (3,87.9) (4,87.9) (5,85.2)};
\addplot[orange!80!black, thick, mark=*, mark size=1.2pt] coordinates {(0,92.6) (1,74.8) (2,80.1) (3,87.1) (4,90.4) (5,91.0)};

\legend{{unedited}, {Fable 5.1}, {Opus 5.5}, {Gemini 4}, {GPT-6 Astra}}
\end{axis}\end{tikzpicture}\caption{Test accuracy against the number of in-context examples, for the unedited model and the updated models (mean over tasks and completed Qwen2.5-0.5B runs; three draws per $k$).}\label{fig:iclcurves}\end{figure}
\IfFileExists{tables/families.tex}{\begin{table*}[t]\centering\small
\resizebox{\textwidth}{!}{\begin{tabular}{llccccccc}
\toprule
Model (editor) & Condition & Antonym & Capital & Past & En$\to$Fr & $2n$ & $n{+}7$ & Mean \\
\midrule
Qwen3.5-0.8B (Astra) & zero-shot, unedited & 5.1$\pm$0.3 & 3.2$\pm$0.5 & 0.0$\pm$0.0 & 0.8$\pm$0.0 & 0.0$\pm$0.0 & 0.0$\pm$0.0 & 1.5 \\
 & 16-shot ICL, unedited & 92.5$\pm$0.1 & 93.8$\pm$1.2 & 96.1$\pm$1.8 & 93.7$\pm$0.4 & 100.0$\pm$0.0 & 97.7$\pm$2.2 & 95.6 \\
 & \textbf{weight-space program} & \textbf{86.0$\pm$4.0} & \textbf{91.5$\pm$2.2} & \textbf{89.5$\pm$2.9} & \textbf{81.7$\pm$10.8} & \textbf{99.1$\pm$1.1} & \textbf{98.1$\pm$3.2} & \textbf{91.0} \\
 & agreement / cos.\ sim. & 79 / 0.71 & 90 / 0.65 & 87 / 0.89 & 80 / 0.71 & 99 / 0.43 & 96 / 0.26 & \\
 & parameters modified & 3.7M & 2K & 3.7M & 3.7M & 1K & 135K & \\ \midrule
Llama-3.2-1B (Gemini) & zero-shot, unedited & 30.5 & 2.6 & 0.0 & 1.6 & 0.0 & 0.0 & 5.8 \\
 & 16-shot ICL, unedited & 98.4 & 99.4 & 91.7 & 96.9 & 83.4 & 80.6 & 91.7 \\
 & \textbf{weight-space program} & \textbf{92.6} & \textbf{100.0} & \textbf{93.5} & \textbf{95.3} & \textbf{100.0} & \textbf{100.0} & \textbf{96.9} \\
 & agreement / cos.\ sim. & 91 / 0.95 & 95 / 0.61 & 88 / 0.68 & 91 / 0.62 & 80 / 0.01 & 80 / 0.00 & \\
 & parameters modified & 55.4M & 16K & 2.6M & 35K & 36.8M & 2.2M & \\ \midrule
Mistral-7B-v0.3 (Gemini) & zero-shot, unedited & 43.7 & 3.5 & 0.0 & 1.6 & 0.7 & 0.7 & 8.4 \\
 & 16-shot ICL, unedited & 98.6 & 100.0 & 100.0 & 100.0 & 100.0 & 100.0 & 99.8 \\
 & \textbf{weight-space program} & \textbf{90.0} & \textbf{99.1} & \textbf{97.6} & \textbf{96.9} & \textbf{100.0} & \textbf{96.5} (turn cap) & \textbf{96.7} \\
 & agreement / cos.\ sim. & 85 / 0.87 & 96 / 0.65 & 97 / 0.82 & 96 / 0.67 & 100 / 0.86 & -- & \\
 & parameters modified & 676K & 33K & 1.4M & 57K & 117.9M & -- & \\
\bottomrule
\end{tabular}}
\caption{Other model families, on the six tasks of Table~\ref{tab:headline}: Gemini 4 Argon (150 turns, one split) on Llama-3.2-1B and Mistral-7B-v0.3, and GPT-6 Astra (three splits, so cells carry a standard deviation) on Qwen3.5-0.8B-Base, a hybrid model whose 24 layers alternate linear and full attention. Columns as in Table~\ref{tab:large}; parameters modified is the median over splits.}
\label{tab:families}
\end{table*}}{}

\subsection{Long-Form Tasks}\label{app:longform}
The grade-school word problems of Appendix~\ref{sec:cot} need answers of a few sentences rather than a word, so three elements of the protocol change for that experiment only. Decoding runs to 256 tokens and stops at the string \texttt{\textbackslash nInput:}, which is where the base model otherwise begins inventing a new problem. An answer counts as correct if the number following ``The answer is'', or else the last number in the output, equals the reference final number; the reference solutions end in ``The answer is $N$.'' and only $N$ is scored, so a model may answer directly or reason first. One call to \texttt{generate} costs one forward pass for up to 64 prompts, instead of one pass per decoding step: at 256 steps the original accounting would spend the whole budget on a single evaluation. The editor is told all three rules in its prompt. The pool is 300 problems drawn with a fixed seed from the GSM8K test split, with calculator annotations removed, split 16 / 30 / 254 as elsewhere; evaluation uses the first 200 test problems.

\section{Why Sixteen Examples}\label{app:shots}
Sixteen pairs is a standard few-shot budget \citep{gao2021making}, and on these tasks it is where the in-context reference stops improving quickly. Table~\ref{tab:shots} varies the number of examples with the test items held fixed. In-context accuracy is \shotIclOne\% with one example, \shotIclFour\% with four, \shotIclEight\% with eight, \shotIclSixteen\% with 16 and \shotIclThirtyTwo\% with 32: with eight examples the reference the updates are measured against would have been \shotGainSixteen\ points lower, and doubling the budget to 32 adds \shotGainThirtyTwo. The gradient baselines move with the count as well, from \shotFtFour\% (Adam), \shotLoraFour\% (LoRA) and \shotRankFour\% (rank-1) at four pairs to \shotFtSixteen\%, \shotLoraSixteen\% and \shotRankSixteen\% at 16 and \shotFtThirtyTwo\%, \shotLoraThirtyTwo\% and \shotRankThirtyTwo\% at 32. Unlike in-context learning, Adam and LoRA are still improving at 32, so more data would narrow the comparison further; with twice the examples they reach what the editors reach with 16. We did not run the editors at other counts; the nearest evidence is the label-free condition, in which the 16 answers are removed and the updates reach \niPairMean\% against \niLab\% with them (Table~\ref{tab:nolabel}).
\IfFileExists{tables/shots.tex}{\begin{table}[t]\centering\small
\resizebox{\columnwidth}{!}{\begin{tabular}{lrrrrrrr}
\toprule
Examples & 0 & 1 & 2 & 4 & 8 & 16 & 32 \\
\midrule
\multicolumn{8}{l}{\textbf{In-context examples, unedited model}} \\
\quad Antonym & 0.0 & 30.8 & 58.5 & 83.2 & 90.4 & 93.8 & 95.5 \\
\quad Capital & 2.7 & 60.2 & 92.4 & 96.7 & 96.8 & 96.5 & 96.9 \\
\quad Past & 0.0 & 67.9 & 89.4 & 92.2 & 95.5 & 93.9 & 94.1 \\
\quad En$\to$Fr & 0.0 & 72.0 & 86.7 & 86.6 & 87.2 & 88.0 & 87.3 \\
\quad $2n$ & 0.0 & 9.4 & 24.0 & 50.3 & 72.4 & 79.8 & 81.0 \\
\quad $n{+}7$ & 0.0 & 14.1 & 29.2 & 56.9 & 68.4 & 83.6 & 82.9 \\
\quad Mean & 0.5 & 42.4 & 63.4 & 77.7 & 85.1 & 89.3 & 89.6 \\
\midrule
\multicolumn{8}{l}{\textbf{Gradient-based, fitted on the same examples (mean over tasks)}} \\
\quad Adam, all weights & -- & -- & -- & 68.1 & 79.5 & 86.3 & 90.5 \\
\quad LoRA & -- & -- & -- & 70.4 & 81.2 & 87.9 & 89.8 \\
\quad Rank-1 FT, matched & -- & -- & -- & 77.3 & 85.2 & 88.4 & 88.6 \\
\bottomrule
\end{tabular}}
\caption{Why 16 examples: zero-shot test accuracy (\%) of Qwen2.5-0.5B against the number of examples, three data splits per task. Every column is scored on the same test items: the last 16 test items of each split are set aside and supply the extra examples for 32. In-context rows average three example orders; gradient rows choose hyper-parameters on the 30 development items, as in Table~\ref{tab:headline}.}
\label{tab:shots}
\end{table}}{}

\section{Pairs Written by the Editors}\label{app:coverage}
The editors are frontier models that know what an antonym or a capital is, and about half of the runs use that knowledge: they type input-output pairs of their own into the kernel and work with them alongside the 16 examples. We extracted every literal pair from the code of \nRunsScanned\ runs of the four standard editors and found such pairs in \nRunsWithPairs: \nPairRunsGemini\ runs of Gemini 4 Argon, \nPairRunsOpus\ of Opus 5.5, \nPairRunsFable\ of Fable 5.1 and \nPairRunsAstra\ of GPT-6 Astra. A run that types pairs adds a median of 72 (from 4 to 356): 170 for Opus 5.5, 80 for Fable 5.1 and 47 for Gemini 4 Argon. Across runs the pairs do not predict accuracy: the \nWithPairs\ completed runs that typed pairs average \accWithPairs\% against \accWithoutPairs\% for the \nWithoutPairs\ that typed none. That comparison is not controlled, and typed pairs are not the only way to bring in task data, since on the arithmetic tasks an editor can generate instances from the rule it infers; the strict-data condition below is the controlled test. To judge by the names the editors give these collections (\texttt{val\_examples}, \texttt{heldout}, \texttt{test\_pairs}, \texttt{extra\_pairs}), most serve as a held-out check of an update, which the protocol gives the editor no other means to run since it is scored on items it never sees, and some enlarge the sample on which the residual is measured. Pooled per task and split (Table~\ref{tab:coverage}), the written pairs are correctly labelled wherever the task defines an answer, and they contain \covMin--\covMax\% of the hidden test inputs (median \covMedian\%) although no editor was shown a test item: the tasks draw on common words, countries and small numbers. A sparse update of three vectors cannot hold a lookup table; for the dense updates we have not ruled memorisation out. ``The same 16 examples'' therefore describes what each method is given, not what it uses. The control in Table~\ref{tab:headline} gives the gradient baselines the \nPairsKept\ editor-written pairs that lie outside the 16 examples, the development set and the test set; \nCellsNoPairs\ of the \nCells\ task-split cells have none, and there the 16-pair run is repeated.
\IfFileExists{tables/coverage.tex}{\begin{table}[t]\centering\small
\resizebox{\columnwidth}{!}{\begin{tabular}{lrrrr}
\toprule
 & Pairs & Test inputs & Labels & Outside \\
Task & written & covered (\%) & correct (\%) & all splits \\
\midrule
Antonym & 219 & 51 (32--80) & 100 & 314 \\
Capital & 93 & 60 (34--95) & 100 & 23 \\
Past & 302 & 97 (94--98) & 100 & 458 \\
En$\to$Fr & 127 & 62 (53--73) & 100 & 83 \\
$2n$ & 29 & 12 (0--18) & 100 & 24 \\
$n{+}7$ & 27 & 15 (3--22) & 100 & 9 \\
\bottomrule
\end{tabular}}
\caption{Task pairs the editors typed into their own code from prior knowledge, pooled over the four editors (mean over three data splits; range in brackets). Pairs written: distinct input--output pairs found in the editors' code. Test inputs covered: the share of the hidden test inputs among them. Labels correct: agreement of the written outputs with the task's reference, on inputs the task contains. Outside all splits: pairs whose input is in neither the 16 examples, the development set nor the test set, summed over splits; these are the extra pairs given to the gradient baselines in Table~\ref{tab:headline}.}
\label{tab:coverage}
\end{table}}{}

\paragraph{The strict-data condition.} To measure what this knowledge is worth we reran three editors on the first split, and Gemini 4 Argon and GPT-6 Astra on two more, with one rule added to the prompt: the 16 pairs are the only instances of the task the editor may use, whether for measuring, for constructing the update or for checking it; it may not type, recall, generate, compute or load any other, with or without answers, while text that is not an instance of the task remains allowed. Everything else is the standard run. All \nStrictRuns\ runs finished, and the extraction used above finds no pair and no input of the task beyond the 16 in the tool calls of any of them; the lines that a pattern check for number generation flagged in \nStrictGenRuns\ arithmetic runs loop over layers, positions, the 16 pairs or the ten digit tokens, or define prompt wordings. Table~\ref{tab:strict} sets each run beside the standard run of the same editor and the baselines on that split. On the first split Gemini 4 Argon reaches \strictGeminiAll\%, Fable 5.1 \strictFableAll\% and GPT-6 Astra \strictAstraAll\%, against \strictFtMin--\strictFtMax\% for the four fine-tuning baselines and \strictIcl\% for ICL: every editor is below fine-tuning on the same 16 pairs, Gemini 4 Argon by two to four points and the other two by six to twelve. The larger losses fall where the standard run drew on data of its own. Fable 5.1 loses 19, 17 and 8 points on capital, translation and past tense, where its standard runs typed 63 to 99 pairs, though only 2 on antonym, where it typed 48, and nothing on $2n$, where it typed none. Gemini 4 Argon is 3 to 10 points lower on each of the five tasks its standard run finished. GPT-6 Astra, which types no pairs, stays within two points on antonym, capital and past tense and falls from 100\% to 37.5\% on $n{+}7$, where its standard run built the update from every input from 0 to 999 with the answers computed in code. One drop has no such explanation: its standard translation run used nothing beyond the 16 pairs and scored 84.5\% where the strict run scored 59.7\%, so two runs of one editor on one split can differ by this much, and with one run per task and editor the per-task differences are indicative only. For Gemini 4 Argon, the strongest of the three under the rule, two further splits give the last block of Table~\ref{tab:strict}: \strictGeminiThree\% over three splits against \strictGeminiThreeFtMin--\strictGeminiThreeFtMax\% for the four baselines and \strictGeminiThreeIcl\% for ICL. Paired over the \strictGeminiThreeCells\ task-split cells it is \strictGeminiThreeBelowMin\ to \strictGeminiThreeBelowMax\ points below the baselines, in no case significantly by the sign-flip test ($p$ from \strictGeminiThreePMin\ to \strictGeminiThreePMax); the bootstrap interval excludes zero only against \strictGeminiThreeTightName, $[\strictGeminiThreeTightLo, \strictGeminiThreeTightHi]$. It is ahead of every baseline on capital, level with the best on translation, and behind every one on antonym, past tense and $2n$. Against its own standard runs it is \strictGeminiThreeSelfDrop\ points lower over the \strictGeminiThreeSelfCells\ cells both finished ($p\strictGeminiThreeSelfP$). GPT-6 Astra on the same three splits reaches \strictAstraThree\%. Paired over the \strictAstraThreeCells\ cells it is \strictAstraThreeBelowMin\ to \strictAstraThreeBelowMax\ points below the baselines, significantly by the sign-flip test against \strictAstraThreeSig\ of the four ($p$ from \strictAstraThreePMin\ to \strictAstraThreePMax, uncorrected), with the tightest interval against \strictAstraThreeTightName, $[\strictAstraThreeTightLo, \strictAstraThreeTightHi]$, and it is \strictAstraThreeSelfDrop\ points below its own standard runs ($p=\strictAstraThreeSelfP$). The loss sits on $n{+}7$ (100\% to 55.1\%), $2n$ (99.3\% to 89.7\%) and translation (85.0\% to 76.2\%); on antonym, capital and past tense it stays within five points. The claim of the main text is therefore the qualified one: the editors match fine-tuning on the 16 pairs when they may add what they know about the task, and fine-tuning given the same additions matches them; held to the 16 pairs, Gemini 4 Argon loses about five points and stays within three of fine-tuning, GPT-6 Astra loses about twelve and falls five to seven below it, and Fable 5.1 falls six to eight points below it on the one split it was run on.
\IfFileExists{tables/strict.tex}{\begin{table}[t]\centering\small
\resizebox{\columnwidth}{!}{\begin{tabular}{lccccccc}
\toprule
Method & Antonym & Capital & Past & En$\to$Fr & $2n$ & $n{+}7$ & Mean \\
\midrule
16-shot ICL, unedited & 94.2 & 96.5 & 93.8 & 86.6 & 80.0 & 80.6 & 88.6 \\
\midrule
\multicolumn{8}{l}{Fine-tuning on the 16 pairs} \\
\quad Adam & 91.6 & 86.8 & 91.9 & 78.3 & 100.0 & 71.5 & 86.7 \\
\quad LoRA & 88.9 & 87.7 & 91.9 & 69.0 & 99.3 & 81.2 & 86.4 \\
\quad Context dist. & 87.9 & 87.7 & 89.5 & 79.8 & 97.9 & 70.1 & 85.5 \\
\quad Rank-1 & 90.0 & 88.6 & 91.9 & 70.5 & 99.3 & 84.7 & 87.5 \\
\midrule
Fable 5.1 & 79.5 & 81.6 & 92.7 & 82.9 & 99.3 & 80.6 & 86.1 \\
\quad the 16 pairs only & \textbf{77.9} & \textbf{62.3} & \textbf{84.7} & \textbf{65.9} & \textbf{99.3} & \textbf{86.8} & \textbf{79.5} \\
\midrule
Gemini 4 Argon & 90.5 & 96.5 & 96.0 & 85.3 & 99.3 & -- & 93.5 \\
\quad the 16 pairs only & \textbf{87.9} & \textbf{88.6} & \textbf{89.5} & \textbf{79.8} & \textbf{89.5} & \textbf{68.1} & \textbf{83.9} \\
\midrule
GPT-6 Astra & 84.7 & 89.5 & 92.7 & 84.5 & 97.9 & 100.0 & 91.6 \\
\quad the 16 pairs only & \textbf{85.8} & \textbf{87.7} & \textbf{91.9} & \textbf{59.7} & \textbf{93.0} & \textbf{37.5} & \textbf{75.9} \\
\midrule
\multicolumn{8}{l}{3 splits} \\
\quad 16-shot ICL, unedited & 93.6 & 97.0 & 93.8 & 86.8 & 80.6 & 83.9 & 89.3 \\
\quad Adam & 90.5 & 88.6 & 93.5 & 79.3 & 99.8 & 67.4 & 86.5 \\
\quad LoRA & 90.0 & 88.0 & 93.3 & 75.2 & 99.1 & 81.7 & 87.9 \\
\quad Context dist. & 90.0 & 88.3 & 91.9 & 79.1 & 99.3 & 79.2 & 88.0 \\
\quad Rank-1 & 90.7 & 87.4 & 92.5 & 73.6 & 98.8 & 86.3 & 88.2 \\
\quad Gemini 4 Argon & 88.9 & 93.6 & 96.8 & 84.2 & 99.1 & 100.0 & 93.8 \\
\quad\quad the 16 pairs only & \textbf{85.8} & \textbf{92.4} & \textbf{89.8} & \textbf{79.8} & \textbf{91.8} & \textbf{75.0} & \textbf{85.8} \\
\quad GPT-6 Astra & 87.7 & 93.6 & 90.1 & 85.0 & 99.3 & 100.0 & 92.6 \\
\quad\quad the 16 pairs only & \textbf{85.4} & \textbf{89.2} & \textbf{90.6} & \textbf{76.2} & \textbf{89.7} & \textbf{55.1} & \textbf{81.0} \\
\bottomrule
\end{tabular}}
\caption{The strict-data condition on Qwen2.5-0.5B, first split (last block: Gemini 4 Argon and GPT-6 Astra on three splits): zero-shot test accuracy (\%). `The 16 pairs only' is the standard run with one rule added to the prompt, that the 16 pairs are the only instances of the task the editor may use, for any purpose; in all 42 runs, the tool calls contain no pair and no task input beyond the 16. The row above each is the same editor's standard run on the same split. A dash marks a run that did not finish; each mean is over the tasks in its row. In the last block each cell averages the splits on which the run finished.}
\label{tab:strict}
\end{table}}{}

\section{Combining Updates and In-Context Examples}\label{sec:stack}

Figure~\ref{fig:iclcurves} in Appendix~\ref{app:tables} adds in-context examples to the models updated under the zero-shot objective. The updates substitute for examples: at $k=0$ the updated models match the unedited model at $k\approx 8$, and at $k=16$ all but the Gemini 4 Argon models match or exceed it. The two do not add up for small $k$: one example lowers every updated model by 5 to 18 points before further examples restore it, because the detector was fitted on the zero-shot format and a prefix of examples displaces the final-token state.

An editor told that every test prompt will carry the same $k$ examples (Table~\ref{tab:hybrid} in Appendix~\ref{app:tables}) does better than either alone: the hybrid update beats the unedited model with $k$ examples, the zero-shot update with them appended (antonym only), and in three of five settings 16-shot ICL, at the cost of larger, less consistently gated updates.

\section{Internalising a Prompt}\label{sec:cot}
The same ability can be pointed at a prompt rather than a mapping. On grade-school word problems \citep{cobbe2021gsm8k}, appending ``Let's think step by step'' to the query \citep{kojima2022zeroshot} raises Qwen2.5-1.5B's accuracy by about eighteen points. We give the editor sixteen worked solutions and the bare evaluation format, say nothing about reasoning or triggers, and ask for an update that makes the bare prompt work. Long-form output needs three protocol changes, used only here (Appendix~\ref{app:longform}): longer decoding, scoring on the final number, and one forward pass per generation call. Table~\ref{tab:cot} reports the results.
\IfFileExists{tables/cot.tex}{\begin{table}[t]\centering\small
\resizebox{\columnwidth}{!}{\begin{tabular}{lccccc}
\toprule
Condition & Astra s0 & Astra s1 & Astra s2 & Gemini s0 & Mean \\
\midrule
Unedited, bare prompt & 41.0 & 41.0 & 40.0 & 41.0 & 40.8 \\
\quad + ``Let's think step by step.'' & 60.0 & 57.0 & 59.0 & 60.0 & 59.0 \\
\quad + 16 worked solutions & 70.0 & 67.5 & 62.0 & 70.0 & 67.4 \\
\midrule
\textbf{Updated, bare prompt} & \textbf{64.0} & \textbf{57.0} & \textbf{58.5} & \textbf{57.0} & \textbf{59.1} \\
\quad + ``Let's think step by step.'' & 59.0 & 58.0 & 57.0 & 59.5 & 58.4 \\
Outputs that reason & 98.5 & 100.0 & 100.0 & 100.0 & 99.6 \\
\midrule
Cos.\ sim.\ with the trigger shift & 0.95 & 0.52 & 0.50 & 0.16 & 0.53 \\
Parameters modified & 13.8M & 13.8M & 3,069 & 64,512 & \\
Perplexity / ARC-Easy & 12.3 / 74 & 12.2 / 77 & 12.2 / 77 & 12.2 / 78 & \\
\bottomrule
\end{tabular}}
\caption{Internalising a prompt rather than a task: GSM8K on Qwen2.5-1.5B (200 held-out problems), scored on the final number. The editor receives 16 worked solutions and the bare \texttt{Input:}/\texttt{Output:} format, and is told nothing about reasoning triggers. Cosine similarity is between the residual shifts of the update and of the trigger phrase at the final prompt token, over the layers the update reaches. The unedited model reaches 40.8\% on the bare prompt because it often reasons unprompted; perplexity of the unedited model is 12.2 and its ARC-Easy accuracy 77\%.}
\label{tab:cot}
\end{table}}{}
\cotSummary{}
The editors reach for the constructions of the word-mapping tasks. Two runs write a rank-one update into a middle-layer MLP output projection, one rewrites the two format-token embeddings, and Gemini 4 Argon builds the gated key-value unit of Figure~\ref{fig:schematic} in the final layer, with a behaviour in place of a mapping. What the first writes is close to the prompt itself: weight-space programming as readable prompt compression.

\section{Two Tasks in One Model}\label{app:twotask}
\IfFileExists{tables/two_task.tex}{\newcommand{\twoIcl}{91.0}\newcommand{\twoIclMin}{89.9}\newcommand{\twoSepAdam}{86.7}\newcommand{\twoSepAdamMin}{83.4}\newcommand{\twoSepLora}{86.4}\newcommand{\twoSepLoraMin}{82.9}%
\newcommand{\twoSepRank}{87.5}\newcommand{\twoSepRankMin}{83.7}\newcommand{\twoSepFable}{86.1}\newcommand{\twoSepFableMin}{84.3}\newcommand{\twoSepOpus}{89.8}\newcommand{\twoSepOpusMin}{87.3}%
\newcommand{\twoSepGemini}{77.9}\newcommand{\twoSepGeminiMin}{62.2}\newcommand{\twoSepAstra}{91.6}\newcommand{\twoSepAstraMin}{87.3}\newcommand{\twoSepFtMin}{86.4}\newcommand{\twoSepFtMax}{87.5}%
\newcommand{\twoJointAdam}{83.7}\newcommand{\twoJointAdamMin}{80.4}\newcommand{\twoJointLora}{83.0}\newcommand{\twoJointLoraMin}{77.9}\newcommand{\twoJointRank}{82.5}\newcommand{\twoJointRankMin}{80.3}%
\newcommand{\twoJointGemini}{84.2}\newcommand{\twoJointGeminiMin}{80.1}\newcommand{\twoJointAstra}{85.0}\newcommand{\twoJointAstraMin}{81.3}\newcommand{\twoJointFtMin}{82.5}\newcommand{\twoJointFtMax}{83.7}%
\newcommand{\twoSeqAdam}{73.8}\newcommand{\twoSeqAdamMin}{53.7}\newcommand{\twoSeqLora}{74.2}\newcommand{\twoSeqLoraMin}{60.1}\newcommand{\twoSeqRank}{70.7}\newcommand{\twoSeqRankMin}{52.6}%
\newcommand{\twoSeqGemini}{72.4}\newcommand{\twoSeqGeminiMin}{47.6}\newcommand{\twoSeqAstra}{77.4}\newcommand{\twoSeqAstraMin}{62.0}\newcommand{\twoSeqFtMin}{70.7}\newcommand{\twoSeqFtMax}{74.2}%
\newcommand{\twoInfAdam}{84.4}\newcommand{\twoInfAdamMin}{78.0}\newcommand{\twoInfLora}{84.0}\newcommand{\twoInfLoraMin}{80.9}\newcommand{\twoInfRank}{83.6}\newcommand{\twoInfRankMin}{78.8}%
\newcommand{\twoInfGemini}{91.7}\newcommand{\twoInfGeminiMin}{88.3}\newcommand{\twoInfFtMin}{83.6}\newcommand{\twoInfFtMax}{84.4}\newcommand{\twoRevAdam}{37.7}\newcommand{\twoRevAdamMin}{1.9}%
\newcommand{\twoRevLora}{42.6}\newcommand{\twoRevLoraMin}{13.9}\newcommand{\twoRevRank}{44.1}\newcommand{\twoRevRankMin}{4.9}\newcommand{\twoRevFtMin}{37.7}\newcommand{\twoRevFtMax}{44.1}%
\newcommand{\twoSumAdam}{44.7}\newcommand{\twoSumAdamMin}{4.7}\newcommand{\twoSumLora}{46.7}\newcommand{\twoSumLoraMin}{6.5}\newcommand{\twoSumRank}{73.6}\newcommand{\twoSumRankMin}{65.1}%
\newcommand{\twoSumFable}{42.4}\newcommand{\twoSumFableMin}{0.0}\newcommand{\twoSumOpus}{47.0}\newcommand{\twoSumOpusMin}{25.9}\newcommand{\twoSumGemini}{31.6}\newcommand{\twoSumGeminiMin}{0.3}%
\newcommand{\twoSumAstra}{56.3}\newcommand{\twoSumAstraMin}{31.1}\newcommand{\twoSumFtMin}{44.7}\newcommand{\twoSumFtMax}{73.6}\newcommand{\twoCostAstra}{9}\newcommand{\twoCostGemini}{53}%
\newcommand{\twoRuns}{15}%
\renewcommand{\dbltopfraction}{0.8}%
\begin{table*}[t]\centering\small
\begin{tabular}{lcccccc}
\toprule
Method & Antonym / capital & Past / $2n$ & En$\to$Fr / $n{+}7$ & Mean & Weaker & PPL \\
\midrule
Unedited, the 32 pairs in context & 93.9 / 93.6 & 91.9 / 94.9 & 84.2 / 87.3 & 91.0 & 89.9 & 17.3 \\
\midrule
\multicolumn{7}{l}{Two single-task models, each scored on its own task} \\
\quad Adam & 91.6 / 86.8 & 91.9 / 100.0 & 78.3 / 71.5 & 86.7 & 83.4 & -- \\
\quad LoRA & 88.9 / 87.7 & 91.9 / 99.3 & 69.0 / 81.2 & 86.4 & 82.9 & -- \\
\quad Rank-1 & 90.0 / 88.6 & 91.9 / 99.3 & 70.5 / 84.7 & 87.5 & 83.7 & -- \\
\quad Fable 5.1 & 79.5 / 81.6 & 92.7 / 99.3 & 82.9 / 80.6 & 86.1 & 84.3 & -- \\
\quad Opus 5.5 & 87.9 / 89.5 & 90.3 / 96.5 & 83.7 / 91.0 & 89.8 & 87.3 & -- \\
\quad Gemini 4 Argon & 90.5 / 96.5 & 96.0 / 99.3 & 85.3 / 0.0 & 77.9 & 62.2 & -- \\
\quad GPT-6 Astra & 84.7 / 89.5 & 92.7 / 97.9 & 84.5 / 100.0 & 91.6 & 87.3 & -- \\
\midrule
\multicolumn{7}{l}{Joint: one update from the 32 mixed pairs} \\
\quad Adam & 83.7 / 77.2 & 94.4 / 100.0 & 69.8 / 77.1 & 83.7 & 80.4 & 17.8 \\
\quad LoRA & 77.4 / 86.0 & 91.1 / 100.0 & 65.1 / 78.5 & 83.0 & 77.9 & 18.3 \\
\quad Rank-1 & 80.0 / 81.6 & 91.9 / 100.0 & 69.0 / 72.2 & 82.5 & 80.3 & 17.4 \\
\quad Gemini 4 Argon & 84.7 / 84.2 & 87.9 / 98.6 & 68.2 / 81.2$^\dagger$ & 84.2 & 80.1 & 17.6 \\
\quad GPT-6 Astra & 77.9 / 86.8 & 83.1 / 88.8 & 82.9 / 90.3 & 85.0 & 81.3 & 17.6 \\
\midrule
\multicolumn{7}{l}{Sequential: first task, then second} \\
\quad Adam & 92.1 / 78.1 & 82.3 / 99.3 & 0.8 / 90.3 & 73.8 & 53.7 & 17.9 \\
\quad LoRA & 76.8 / 80.7 & 91.9 / 100.0 & 11.6 / 84.0 & 74.2 & 60.1 & 19.8 \\
\quad Rank-1 & 78.4 / 78.1 & 79.0 / 98.6 & 0.8 / 89.6 & 70.7 & 52.6 & 17.7 \\
\quad Gemini 4 Argon & 0.0 / 96.5 & 87.1 / 100.0 & 55.8 / 95.1$^\dagger$ & 72.4 & 47.6 & $>10^{12}$ \\
\quad GPT-6 Astra & 67.4 / 83.3 & 37.9 / 95.1 & 80.6 / 100.0 & 77.4 & 62.0 & 17.5 \\
\midrule
\multicolumn{7}{l}{Sequential, with the first task in view at the second stage} \\
\quad Adam & 88.4 / 83.3 & 86.3 / 100.0 & 64.3 / 84.0 & 84.4 & 78.0 & 18.0 \\
\quad LoRA & 84.2 / 76.3 & 93.5 / 100.0 & 72.9 / 77.1 & 84.0 & 80.9 & 19.8 \\
\quad Rank-1 & 87.4 / 76.3 & 87.9 / 99.3 & 72.1 / 78.5 & 83.6 & 78.8 & 17.5 \\
\quad Gemini 4 Argon & 90.0 / 86.0 & 96.0 / 93.7 & 85.3 / 99.3 & 91.7 & 88.3 & 17.8 \\
\midrule
\multicolumn{7}{l}{Sequential: second task, then first} \\
\quad Adam & 92.1 / 0.0 & 91.9 / 0.0 & 36.4 / 5.6 & 37.7 & 1.9 & 19.3 \\
\quad LoRA & 83.2 / 19.3 & 91.9 / 0.0 & 22.5 / 38.9 & 42.6 & 13.9 & 21.9 \\
\quad Rank-1 & 88.4 / 3.5 & 94.4 / 0.0 & 67.4 / 11.1 & 44.1 & 4.9 & 17.8 \\
\midrule
\multicolumn{7}{l}{Sum of the two single-task updates} \\
\quad Adam & 92.1 / 0.9 & 91.1 / 12.6 & 0.8 / 70.8 & 44.7 & 4.7 & 17.9 \\
\quad LoRA & 87.9 / 0.0 & 90.3 / 0.0 & 19.4 / 82.6 & 46.7 & 6.5 & 22.5 \\
\quad Rank-1 & 72.1 / 88.6 & 79.0 / 96.5 & 44.2 / 61.1 & 73.6 & 65.1 & 18.1 \\
\quad Fable 5.1 & 79.5 / 0.0 & 92.7 / 0.0 & 82.2 / 0.0 & 42.4 & 0.0 & 17.6 \\
\quad Opus 5.5 & 24.2 / 3.5 & 74.2 / 96.5 & 83.7 / 0.0 & 47.0 & 25.9 & 17.4 \\
\quad Gemini 4 Argon & 0.0 / 96.5 & 0.8 / 7.0 & 85.3 / 0.0 & 31.6 & 0.3 & 22.1 \\
\quad GPT-6 Astra & 72.1 / 4.4 & 89.5 / 88.8 & 82.9 / 0.0 & 56.3 & 31.1 & 18.1 \\
\bottomrule
\end{tabular}
\caption{Two tasks in one model (Qwen2.5-0.5B, first split, one run per cell). A cell gives the test accuracy (\%) on the first / second task of the pair; queries of both tasks use the one format and nothing marks the task. Mean: over the six task cells. Weaker: the lower accuracy of each pair, averaged over the pairs. PPL: the highest perplexity among the three models (17.3 unedited). The editors' sequential runs receive the second task on a model that carries their own program for the first, and are not told. In the rows with the first task in view the editor is told of the first task and given its 16 pairs, and the second stage of fine-tuning trains on the 32 pairs of both tasks. $^\dagger$The run reached the 150-turn limit and is scored as it stood. Gemini 4 Argon's single-task $n{+}7$ run had failed, which lowers its two-model and summed rows.}
\label{tab:twotask}
\end{table*}
}{}
\paragraph{Summary.} This appendix asks whether one model can hold two tasks when nothing in the prompt says which one is asked. Written together, editing and fine-tuning are level (\twoJointGemini\% and \twoJointAstra\% against \twoJointFtMin--\twoJointFtMax\%, averaged over both tasks). Written one after the other, both forget part of the first task (\twoSeqGemini\% and \twoSeqAstra\% against \twoSeqFtMin--\twoSeqFtMax\%), and two separate updates added together mostly break each other. The editors' programs collide in the way Section~\ref{sec:mechanism} predicts: a program is switched on by the prompt format, which the two tasks share, and not by the task. An editor that is told of the first task reads the earlier program and keeps the two apart (\twoInfGemini\%, against \twoInfFtMin--\twoInfFtMax\% for fine-tuning on the examples of both tasks). Each number is one run.

\paragraph{Setup.} The main experiments write one task into one model. Here two tasks share a model and a prompt format: a query is rendered as before and nothing marks its task, so the model must tell the two apart from the input alone. We pair antonym with capital, past tense with $2n$ and English$\to$French with $n{+}7$ on the first split of Qwen2.5-0.5B and score each task on its own test items (Table~\ref{tab:twotask}). Three routes to one model are compared. Joint: fine-tuning on the 32 pairs, or an editor given the 32 pairs and told that the list mixes two tasks, 16 pairs of each, unmarked. Sequential: fine-tuning on one task starting from the model fine-tuned on the other, or an editor given the second task on a model that carries its own program for the first, which it is not told. Summed: the two single-task updates of the main study added together. The fine-tuning baselines keep their grids, and each stage is selected on the development items of the data it is trained on. The editors are Gemini 4 Argon with 150 turns and GPT-6 Astra; the single-task programs of all four editors are summed. The references are the unedited model with the 32 pairs in context (\twoIcl\%) and each method's two single-task models.

\paragraph{Joint.} One update for two tasks costs a few points, and the editors are level with fine-tuning: \twoJointGemini\% for Gemini 4 Argon and \twoJointAstra\% for GPT-6 Astra against \twoJointFtMin--\twoJointFtMax\% for the three baselines, where two separate fine-tuned models reach \twoSepFtMin--\twoSepFtMax\%. Both editors separate the tasks by the type of the input. GPT-6 Astra writes a rank-two change to one MLP matrix on the first pair and one matrix per task on the second; on the third it rewrites the two format-token embeddings and adds rank-8 corrections in two layers. On the first pair Gemini 4 Argon fits a linear read of the input type at the final prompt token and with it switches one of two task vectors on in each of layers 12 to 23; on the second it adds a digit-and-carry circuit for $2n$. Its three joint updates change 52M to 105M parameters.

\paragraph{Sequential.} Written in turn, both kinds of update forget, in different places. Fine-tuning on $n{+}7$ erases translation (0.8--11.6\%), where the editors keep 80.6\% and 55.8\%; the editors lose antonym (67.4\% and 0\%) and GPT-6 Astra loses past tense (37.9\%), which fine-tuning mostly keeps. Over the three pairs the editors reach \twoSeqGemini\% and \twoSeqAstra\% against \twoSeqFtMin--\twoSeqFtMax\%. In the reverse order, run for fine-tuning only, the task trained first falls to 0--39\% (\twoRevFtMin--\twoRevFtMax\% overall). The editors' forgetting is interference, not overwriting: in four of their six sequential runs every tensor of the first program is identical after the second edit, and the first task still falls, by 4 to 55 points for GPT-6 Astra and from 90.5\% to 0\% for Gemini 4 Argon. In that run the second edit copies the embeddings of \texttt{Country} and \texttt{Capital} into the two format tokens, as Gemini 4 Argon's single-task capital program does. The antonym program is intact in the weights and scores 0\%, since every query is now processed as a country-capital prompt.

\paragraph{Sequential, with the first task in view.} The sequential editors above are not told that the model has been edited. We repeated Gemini 4 Argon's three sequential runs with a prompt that says so: the model already carries an edit for an earlier task, these are the 16 pairs it was built from, and both tasks are scored. The prompt does not say where the edit is. The matching baseline is fine-tuning with replay, whose second stage starts from the first task's fine-tune and trains on the 32 pairs of both tasks. With the first task in view the editor keeps it, at 90.0\%, 96.0\% and 85.3\% against 90.5\%, 96.0\% and 85.3\% for the first program alone, and reaches \twoInfGemini\% over the three pairs, where its runs that were not told reach \twoSeqGemini\% and its joint runs \twoJointGemini\%. Replay returns fine-tuning to its joint level, \twoInfFtMin--\twoInfFtMax\%. The editor's lead over the best replay baseline is uneven: 17 points on translation with $n{+}7$ (92.3\% against 75.3\%), where its translation program was also the stronger first stage (85.3\% against 69.0--78.3\%), two points on antonym with capital (88.0\% against 85.9\%), and on past tense with $2n$ LoRA with replay is two points ahead (96.8\% against 94.8\%). In every run the editor first finds the earlier edit, by comparing the weights with the original checkpoint, which the kernel can read, and then tests what it does on inputs of the new task. On the first two pairs it finds the interference described above, keys that open on the format for inputs of either task, and repairs it inside the earlier program. It rewrites the keys of 12 of the 13 antonym units and the gate rows of all 22 past-tense units so that they stay closed on inputs of the new task, leaves every value as it was, and writes the new task into unused MLP units, 16 for capital and 446 for $2n$, nearly all of them keyed on the type of the input; on the second pair it also rewrites the earlier changes to the attention key projections of layers 14 to 23. On the third pair every entry of the translation program is unchanged, and six new MLP units are added: four in layers 0 and 1 move the states of the tokens that follow a number onto those of \texttt{+7 =}, which hands the query to the model's own addition, and two in layer 15 correct its carry. Replay needs the first task's pairs and so does the editor; what the editor adds is that it reads the earlier update and patches it. No run reached the turn limit (101 to 132 turns). The kernel was not confined (Appendix~\ref{app:open}): a scan of the three transcripts finds no access to the task data, to results or to other runs, and no gradient loop, and as in the main study the editor checks its update on pairs it writes itself (Appendix~\ref{app:coverage}).

\paragraph{Summed.} Single-task updates do not add up. The sum of two Adam or LoRA fine-tunes keeps one task and loses most of the other (\twoSumAdam\% and \twoSumLora\%), and so do the stacked programs: of the eleven stacks whose two programs both work, nine leave one task below 8\%. Only the rank-1 fine-tunes add up in part, to \twoSumRank\%. This agrees with the format detector of Section~\ref{sec:mechanism}: a key fitted on the shared format fires for both tasks.

\paragraph{Caveats.} There is one split and one run per cell, sequential editing was run in one order, and the first task was put in view for one editor. Two Gemini 4 Argon runs, both on translation with $n{+}7$, reached the turn limit and are scored as they stood. The sequential one left a single gate weight of about 2\,600 in the last layer, which takes perplexity above $10^{12}$ and ARC-Easy from 66.5\% to 33.5\%; every other editor run leaves perplexity between 17.35 and 17.83, against 17.34 unedited.

\section{Open-Weight Editors}\label{app:open}
\newcommand{\openIcl}{88.6}\newcommand{\openAdam}{86.7}\newcommand{\openFrontMin}{86.1}\newcommand{\openFrontMax}{91.6}\newcommand{\openRuns}{12}\newcommand{\openDeepseekMean}{73.9}%
\newcommand{\openDeepseekDone}{3}\newcommand{\openDeepseekCost}{2.62}\newcommand{\openGlmMean}{5.2}\newcommand{\openGlmDone}{5}\newcommand{\openGlmCost}{7.77}\newcommand{\openCost}{62}%
\newcommand{\openLeakAcc}{86.3}\newcommand{\openLeakCost}{3.63}\newcommand{\openUntied}{81.9}\newcommand{\openTied}{81.9}\newcommand{\openTiedPpl}{17.5}%
\begin{table*}[t]\centering\small
\begin{tabular}{lccccccccc}
\toprule
Editor & Antonym & Capital & Past & En$\to$Fr & $2n$ & $n{+}7$ & Mean & Done & Turns \\
\midrule
16-shot ICL, unedited & 94.2 & 96.5 & 93.8 & 86.6 & 80.0 & 80.6 & 88.6 & & \\
Adam, all weights & 91.6 & 86.8 & 91.9 & 78.3 & 100.0 & 71.5 & 86.7 & & \\
\midrule
\multicolumn{10}{l}{Editors of the main study} \\
\quad Fable 5.1 & 79.5 & 81.6 & 92.7 & 82.9 & 99.3 & 80.6 & 86.1 & 6/6 & 51 \\
\quad Opus 5.5 & 87.9 & 89.5 & 90.3 & 83.7 & 96.5 & 91.0 & 89.8 & 6/6 & 64 \\
\quad Gemini 4 Argon & 90.5 & 96.5 & 96.0 & 85.3 & 99.3 & 0.0$^{\dagger}$ & 77.9 & 5/6 & 91 \\
\quad GPT-6 Astra & 84.7 & 89.5 & 92.7 & 84.5 & 97.9 & 100.0 & 91.6 & 6/6 & 16 \\
\midrule
\multicolumn{10}{l}{Open-weight editors} \\
\quad DeepSeek V4.1 Flash & 76.8$^{\dagger}$ & 100.0 & 84.7$^{\dagger\S}$ & 83.7 & 97.9$^{\dagger}$ & 0.0$^{\ddagger}$ & 73.9 & 3/6 & 53 \\
\qquad perplexity & 17.4 & $>10^{35}$ & 134 & 17.3 & $>10^{11}$ & -- &  & & \\
\qquad ARC-Easy & 65.0 & 26.0 & 50.0 & 66.5 & 25.5 & -- &  & & \\
\quad GLM-5.2 & 2.6 & 28.1 & 0.0$^{\ddagger}$ & 0.0 & 0.0 & 0.7$^{\dagger}$ & 5.2 & 5/6 & 56 \\
\qquad perplexity & 19.0 & 18.8 & -- & $>10^{10}$ & $>10^{6}$ & 17.8 &  & & \\
\qquad ARC-Easy & 59.5 & 62.5 & -- & 36.5 & 34.0 & 68.0 &  & & \\
\bottomrule
\end{tabular}
\caption{Open-weight editors on Qwen2.5-0.5B (first split, one run per cell): test accuracy (\%), beside the references and the four editors of the main study on the same split (Gemini 4 Argon with 150 turns). Under each open-weight editor: WikiText-2 perplexity and ARC-Easy accuracy (\%) of the updated model (17.3 and 66.5 unedited). Done: runs that ended with a final message. Turns: mean per run. $^\dagger$The run reached the turn limit or, for the open-weight editors, the cost cap, and is scored as it stood. $^\ddagger$The update added a parameter the model does not have and could not be stored; it is scored as no update. $^\S$Fitted with a hand-written gradient-descent loop, which the task forbids.}
\label{tab:open}
\end{table*}

The four editors of the main study are proprietary. For this reason, we gave the standard condition to two open-weight models, DeepSeek V4.1 Flash and GLM-5.2: the six tasks on the first split of Qwen2.5-0.5B, one run per task, with the prompt, the tool, the 50 forward passes and the 80 turns of the Fable 5.1, Opus 5.5 and GPT-6 Astra runs (Table~\ref{tab:open}). The two are reasoning models with open weights under the MIT licence \citep{deepseek2026v41flash,zai2026glm52}. DeepSeek V4.1 Flash has 552B backbone parameters, and GLM-5.2 was Z.ai's flagship model at its release. Both are served by Vertex AI through a chat-completions interface (\texttt{deepseek-v4.1-flash-maas} and \texttt{glm-5.2-maas}), with reasoning requested at the highest setting the endpoint accepts and the \texttt{python} function as the only tool; for DeepSeek the one-sentence system prompt opens the first user message, as the provider advises for function calling. Reasoning text came back in 652 of the 655 model turns of the twelve runs. Two things differ from the main study. A cost cap at list prices, \$5 per run for DeepSeek V4.1 Flash and \$10 for GLM-5.2, ended two runs before the turn limit. The kernel was confined, as presented next.

\paragraph{Confinement.} The first two runs of this experiment, both on antonym, used the unconfined kernel of the main study. From its 25th turn DeepSeek V4.1 Flash imported the harness's task module, printed the file that defines the tasks, rebuilt the split with its test items, read the source and the logs of the scripted task-vector baseline, and checked its update on the held-out items; it stood at \openLeakAcc\% when the turn limit arrived. That run is set aside and appears in no table. GLM-5.2's run opens no file and is kept. The other eleven runs, among them a second antonym run of DeepSeek V4.1 Flash, were confined before their first turn. The task module and the development and test items are deleted from the process. \mbox{Landlock}, the Linux access-control module, limits the files the process can reach to the system directories, the model weights, the helper source and the run's own folder. An audit hook refuses, in code the editor submits, file access outside a scratch folder and those read-only locations, new processes and threads, and network calls. In the eleven confined runs no tool call attempts to reach the task data, the results or another run; the hook refused one call, a write to \texttt{/tmp}.

\paragraph{GLM-5.2.} GLM-5.2 produces no working update (\openGlmMean\% over the six tasks). It fits the examples and reports success, and the fit barely transfers. On antonym it rewrites one MLP unit of the last layer, as the editors of the main study often do, but the direction the unit writes raises the 16 training answers and carries no task vector: 93.8\% on the examples and 2.6\% on the test items. Its capital update stores the 16 pairs as keys and values in three MLP output matrices (87.5\% and 28.1\%), and its translation update fits the last MLP output matrix to them (100\% and 0\%, with perplexity above $10^{10}$). The past-tense update adds a bias vector to a projection that has none. The harness stores an update as changes to the model's own parameters, so nothing is stored and the run scores 0\%; what the model with the added bias would score on the test items was not measured. On the two arithmetic tasks no update works even on the examples (0\% and 37.5\%), and the cost cap ended the second run.

\paragraph{DeepSeek V4.1 Flash.} DeepSeek V4.1 Flash scores above 75\% on five tasks, and one of its six updates is sound. That one is translation: it copies the embeddings of \texttt{English} and \texttt{French} into the format tokens \texttt{Input} and \texttt{Output}, the format rewrite that the frontier editors also use (Table~\ref{tab:editors}), and scores 83.7\% against their 82.9--85.3\% with perplexity unchanged. Its antonym update, a dense change to the MLP output matrix of layer 15, leaves the model intact and stands below all four of theirs, at 76.8\%, when the turn limit arrives. On $n{+}7$ it rewrites three format-token embeddings so that the prompt reads as \texttt{Input = $x$ +7 =}, but first replaces the output head, which shares its matrix with the embeddings, by a separate copy. That is a new parameter, so no update is stored and the run scores 0\%. Rebuilt from the transcript, the model scores \openUntied\% on the test items, and \openTied\% with the head left as it was (perplexity \openTiedPpl): the step that voided the run was not needed. The other three updates are accurate, and each either destroys the model or breaks a rule. For capital it types 235 countries with their capitals and stores them in 795 units of the last MLP: 100\%, with perplexity above $10^{35}$ and ARC-Easy at chance, in a final report that calls the side effects limited. For $2n$ it clears about half of the residual dimensions in every matrix of the model and writes a digit-by-digit doubling circuit into them (97.9\%, perplexity above $10^{11}$, ARC-Easy at chance); the cost cap ended that run at turn 60. For past tense it fits the last MLP output matrix to about 200 verb pairs, most of them its own (84.7\%, perplexity 134, ARC-Easy 50.0\%), and it finds the targets of that fit by gradient descent written out by hand, with analytic gradients and a line search. The prompt forbids training loops; the harness's filter matches autograd and optimiser calls (Appendix~\ref{app:setup}) and does not see a loop written this way. A pattern check for such loops finds this run, one call in GLM-5.2's $n{+}7$ run, and nothing in the transcripts of the four frontier editors.

\paragraph{Summary and caveats.} Scored as in the main study, DeepSeek V4.1 Flash averages \openDeepseekMean\% and GLM-5.2 \openGlmMean\%, against \openFrontMin--\openFrontMax\% for the three frontier editors that finished all six runs, \openAdam\% for fine-tuning and \openIcl\% for ICL; \openDeepseekDone\ of DeepSeek V4.1 Flash's six runs and \openGlmDone\ of GLM-5.2's ended with a final message. The parts of the frontier editors' work are within reach of an open-weight model: the format rewrite, the closed-form fit, a circuit written by hand. Neither model reproduces the combination the main study reports, an accurate update that is stored in the model's own weights, made within the rules and the budget, and harmless to the rest of the model. DeepSeek V4.1 Flash manages it once in six runs and GLM-5.2 not at all. The frontier editors' completed runs on this split leave perplexity between 17.3 and 18.1, with one exception, 22.1 for Gemini 4 Argon on $2n$ (one Fable 5.1 run was evaluated without the measure). Unconfined, DeepSeek V4.1 Flash read the test set. The evidence is one run per task on one split for two models, with cost caps low enough to end two runs. The \openRuns\ runs of Table~\ref{tab:open} cost \$\openCost\ at list prices, \$\openDeepseekCost\ per run for DeepSeek V4.1 Flash and \$\openGlmCost\ for GLM-5.2, and the run set aside \$\openLeakCost.

\section{Strategy Annotation}\label{app:taxonomy}
Each run's final message and the summaries of its reasoning were submitted to Claude Sonnet 5 with the instruction to classify the editor's approach under a fixed schema with structured output: primary strategy (task-vector injection; format-token rewrite; attention edit; embedding edit; steering unit; bias injection; other); whether the update is gated by a context detector; whether the editor estimated the few-shot-minus-zero-shot activation difference; whether it used the terms ``task vector'' or ``function vector''; whether it measured general language-modelling degradation; whether it abandoned a first approach; the layers modified; and a one-sentence summary. Annotations were not validated by human raters.

\subsection{Transcript Excerpt}
The following is the fifth tool call of the seed-0 antonym run of Fable 5.1 on Qwen2.5-0.5B, after the editor had inspected the helper code and tokenisation. It defines a single-pass capture of logits and residual states and compares the zero-shot predictions with the references; the output (abridged) shows the model copying inputs, which the editor then used to motivate measuring the few-shot minus zero-shot difference.

{\footnotesize
\begin{verbatim}
@torch.no_grad()
def fwd(prompts, keep_all_positions=False):
    outs = {"logits": [], "hid": [], "all": []}
    for i in range(0, len(prompts), 64):
        batch = prompts[i:i+64]
        enc = tok(batch, return_tensors="pt",
                  padding=True).to(model.device)
        o = model(**enc, output_hidden_states=True)
        last = o.logits[:, -1, :].float().cpu()
        outs["logits"].append(last)
        hid = [h[:, -1, :].float().cpu()
               for h in o.hidden_states]
        outs["hid"].append(torch.stack(hid, 1))
    ...
prompts = [make_prompt(x)
           for x, _ in train_examples]
r0 = fwd(prompts)
for (x, y), lg in zip(train_examples,
                      r0["logits"]):
    pred = tok.decode(lg.argmax())
    print(f"{x:12s} -> {y:12s} pred={pred!r}")

>>> birth        -> death        pred=' '
>>> rare         -> common       pred=' rare'
>>> death        -> life         pred=' '
>>> teach        -> learn        pred=' teach'
>>> [budget: 1/50 passes used; 1.6s]
\end{verbatim}}

\section{Incomplete Runs}\label{app:failures}
Not every run ends with an update in place. With the 80-turn budget, Gemini 4 Argon completed one run in six: in four it applied, tested, and reverted candidate updates without committing before the budget was exhausted, and in one ($n\to 2n$) the budget expired during an ungated 24-layer injection that solved the task (100\% against 80\% for ICL) while raising WikiText-2 perplexity from 17.3 to 26.3 and reducing ARC-Easy accuracy from 66.5\% to 45.5\%. Two Fable 5.1 runs on $n\to n{+}7$ likewise exhausted their budget, one leaving a model with perplexity above 800; a 150-turn Gemini run on the same task (seed 1) expired mid-edit with perplexity at 26.9; and four Opus 5.5 runs at seeds 1 and 2 (two on $n\to n{+}7$, one each on past tense and $n\to 2n$) hit the cap, two of them with a working update in place and two with none. Two Gemini runs were terminated by the provider's content filter in mid-run, once at turn 57 with an update in place (88.4\% on antonym) and once at turn 129 with none. All such runs are excluded from aggregate figures and counted in Table~\ref{tab:editors}. A further caveat concerns the budget: Gemini 4 Argon re-implemented the forward pass as tensor algebra on the CPU (maximum logit error $5.7\times 10^{-5}$) and evaluated candidate updates through it, consuming one to six of its fifty passes in 80-turn runs. A budget counted in forward passes means different things for different editors. Two further 150-turn Gemini runs (seed 2, $n\to 2n$ and $n\to n{+}7$) failed to start because the shared GPU was full and were relaunched. In the label-free condition four 150-turn Gemini runs hit the cap, two on translation and two on $n\to n{+}7$, one of each with a working update in place and one with none; Table~\ref{tab:nolabel} leaves them out. Table~\ref{tab:failures} lists every run that did not complete, except the two of the two-task study and those of the open-weight editors, which Appendices~\ref{app:twotask} and~\ref{app:open} report.
\IfFileExists{tables/failures.tex}{\begin{table*}[htbp]\centering\small
\resizebox{\textwidth}{!}{\begin{tabular}{llllrrrrr}
\toprule
Editor & Model & Task & Seed & Ending & Turns & Acc.\ left & ICL & PPL after \\
\midrule
Fable 5.1 & Qwen2.5-0.5B & $n{+}7$ & 1 & turn cap & 80 & 99.3 & 84.0 & 873.5 \\
Fable 5.1 & Qwen2.5-0.5B & $n{+}7$ & 2 & turn cap & 80 & 100.0 & 87.0 & 24.7 \\
Opus 5.5 & Qwen2.5-0.5B & $n{+}7$ & 1 & turn cap & 80 & 88.2 & 84.0 & 17.3 \\
Opus 5.5 & Qwen2.5-0.5B & $n{+}7$ & 2 & turn cap & 80 & 0.0 & 87.0 & 17.4 \\
Opus 5.5 & Qwen2.5-0.5B & Past & 1 & turn cap & 80 & 94.4 & 91.7 & 17.3 \\
Opus 5.5 & Qwen2.5-0.5B & $2n$ & 2 & turn cap & 80 & 0.0 & 77.4 & 17.3 \\
Gemini 4 (80 turns) & Qwen2.5-0.5B & $n{+}7$ & 0 & turn cap & 80 & 0.0 & 80.6 & 17.3 \\
Gemini 4 (80 turns) & Qwen2.5-0.5B & Antonym & 0 & provider block & 57 & 88.4 & 94.2 & 17.4 \\
Gemini 4 (80 turns) & Qwen2.5-0.5B & Capital & 0 & turn cap & 80 & 2.6 & 96.5 & 17.3 \\
Gemini 4 (80 turns) & Qwen2.5-0.5B & En$\to$Fr & 0 & turn cap & 80 & 0.0 & 86.6 & 17.3 \\
Gemini 4 (80 turns) & Qwen2.5-0.5B & Past & 0 & turn cap & 80 & 0.0 & 93.8 & 17.3 \\
Gemini 4 (80 turns) & Qwen2.5-0.5B & $2n$ & 0 & turn cap & 80 & 100.0 & 80.0 & 26.3 \\
Gemini 4 (150 turns) & Qwen2.5-0.5B & $n{+}7$ & 0 & provider block & 129 & 0.0 & 80.6 & 17.3 \\
Gemini 4 (150 turns) & Qwen2.5-0.5B & $n{+}7$ & 1 & turn cap & 150 & 89.6 & 84.0 & 26.9 \\
Gemini 4 (150 turns) & Qwen2.5-14B & last letter & 0 & turn cap & 150 & 99.3 & 90.8 & 10196.2 \\
Gemini 4 (150 turns) & Qwen2.5-7B & camelCase & 0 & turn cap & 150 & 90.9 & 99.8 & 7.8 \\
Gemini 4 (150 turns) & Mistral-7B-v0.3 & $n{+}7$ & 0 & turn cap & 150 & 96.5 & 100.0 & 8.1 \\
Gemini 4 (150 turns, no labels) & Qwen2.5-0.5B & $n{+}7$ & 0 & turn cap & 150 & 0.0 & 80.6 & 17.3 \\
Gemini 4 (150 turns, no labels) & Qwen2.5-0.5B & $n{+}7$ & 2 & turn cap & 150 & 98.6 & 87.0 & 17.4 \\
Gemini 4 (150 turns, no labels) & Qwen2.5-0.5B & En$\to$Fr & 1 & turn cap & 150 & 0.0 & 86.6 & 17.3 \\
Gemini 4 (150 turns, no labels) & Qwen2.5-0.5B & En$\to$Fr & 2 & turn cap & 150 & 79.1 & 87.3 & 17.8 \\
\bottomrule
\end{tabular}}
\caption{Editor runs that did not complete, and the state they left behind. Ending: turn cap = the turn budget expired without a final message; provider block = the API refused a request mid-run. Acc.\ left: zero-shot test accuracy of the model as left, which may contain a partial update or none; ICL: 16-shot accuracy of the unedited model; PPL after: WikiText-2 perplexity of the model as left where measured (unedited: 17.3 on Qwen2.5-0.5B, 7.8 on 7B, 4.8 on 14B, 7.7 on Mistral-7B). These runs are excluded from every mean in the paper.}
\label{tab:failures}
\end{table*}}{}

\section{Cost}\label{app:cost}
Table~\ref{tab:cost} reports usage and cost per condition. The Fable 5.1, Opus 5.5 and Gemini 4 Argon amounts are the totals billed by the providers for the study (\$\spendFable; \$\spendOpus, of which \$30 was billed for the first six Opus runs and the twelve later runs are counted at the same effective rate; and about \texteuro300, converted to \$\spendGemini\ at 1.08 USD per EUR), distributed over runs in proportion to each run's token usage; the Opus and Gemini totals are approximate and cover the runs completed by 8 November 2026. The strict-data runs of Fable 5.1 and Gemini 4 Argon came after these bills and are counted at each editor's effective billed rate (\$\spendLate\ together). The GPT-6 Astra amount (\$\spendAstra) is computed from the token counts the API reports at the rates in force during the study. Two features determine cost. First, prompt caching: the task prompt and the growing transcript are re-sent every turn, and 81--98\% of input tokens, depending on the condition, are served from the cache at a tenth of the input price or less, so the output and reasoning tokens dominate. Second, context length: Gemini 4 Argon's transcripts exceed 150\,000 tokens by turn 50 because its tool outputs are long, so its token usage per run is two to four times that of the Claude editors, although at the billed rates its cost per run is close to theirs. Evaluation and baselines add approximately 13 GPU-hours in total. The total editor usage for every run reported in this paper, including the runs that did not complete, was approximately \$\spend. The \twoRuns\ two-task runs of Appendix~\ref{app:twotask} came later still and are in neither the table nor this total: \$\twoCostAstra\ for GPT-6 Astra from its token counts and about \$\twoCostGemini\ for Gemini 4 Argon at its effective billed rate. Nor are the open-weight editor runs of Appendix~\ref{app:open}, \$\openCost\ at list prices.

\end{document}